\documentclass{article}

\usepackage[preprint]{neurips_2024}

\usepackage[utf8]{inputenc} 
\usepackage[T1]{fontenc}    
\usepackage{hyperref}       
\usepackage{url}            
\usepackage{booktabs}       
\usepackage{amsfonts}       
\usepackage{nicefrac}       
\usepackage{microtype}      
\usepackage{xcolor}         
\usepackage{graphicx}
\usepackage{multirow}
\usepackage{amsmath,amssymb}
\usepackage{amsthm}
\usepackage{mathrsfs}
\usepackage{tabularx}
\usepackage{tabulary}
\usepackage{threeparttable} 
\usepackage{comment}
\usepackage{adjustbox}
\usepackage{array}
\usepackage{adjustbox}
\usepackage{color}
\usepackage{mathrsfs}
\usepackage{times}
\usepackage{epsfig}
\usepackage{multicol}
\usepackage{makecell}
\usepackage{enumitem}
\usepackage{siunitx}
\usepackage{colortbl}
\usepackage{wrapfig}
\usepackage{todonotes}
\usepackage{wrapfig}  
\usepackage{subcaption} 

\newcommand\numberthis{\addtocounter{equation}{1}\tag{\theequation}}

\usepackage{comment}

\title{Enhancing Accuracy and Parameter-Efficiency of Neural Representations for Network Parameterization}

\author{%
  Hongjun Choi \\
  Lawrence Livermore National Laboratory\\
  \texttt{choi22@llnl.gov} \\
  \And
  Jayaraman J. Thiagarajan \\
  Lawrence Livermore National Laboratory\\
  \texttt{jjayaram@llnl.gov} \\
  \And
  Ruben Glatt \\
  Lawrence Livermore National Laboratory\\
  \texttt{glatt1@llnl.gov} \\
  \And
  Shusen Liu \\
  Lawrence Livermore National Laboratory\\
  \texttt{liu42@llnl.gov} \\
}

\begin{document}
\maketitle

\begin{abstract}

In this work, we investigate the fundamental trade-off regarding accuracy and parameter efficiency in the parameterization of neural network weights using predictor networks. We present a surprising finding that, when recovering the original model accuracy is the sole objective, it can be achieved effectively through the weight reconstruction objective alone. Additionally, we explore the underlying factors for improving weight reconstruction under parameter-efficiency constraints, and propose a novel training scheme that decouples the reconstruction objective from auxiliary objectives such as knowledge distillation that leads to significant improvements compared to state-of-the-art approaches. Finally, these results pave way for more practical scenarios, where one needs to achieve improvements on both model accuracy and predictor network parameter-efficiency simultaneously.

\end{abstract}
\section{Introduction}
\label{sec:introduction}
%

Recently, neural network weight space exploration and manipulation have gained an increase in popularity as alternative avenues for traditional model training and fine-tuning.
Examples of these methods range from weight manipulation strategies such as weight merging to improve model performance without fine-tuning \citep{matena2022merging} to weight generation approaches that directly predict network parameters \citep{ashkenazi2022nern, soro2024diffusion}.
One of these approaches uses implicit neural representations (INRs) to directly represent the weights of a pre-trained convolutional neural network (CNN), where the coordinates of the kernel or layer indices in the INR are mapped to the corresponding weights in the CNN \citep{ashkenazi2022nern}. 
Despite the interesting formulation of the problem, the approach provides limited practical benefits, as the reconstructed weights invariably exhibit worse performance compared to the original model.



In this work, we are
exploring the various trade-offs between predictor networks and reconstructed models as well as the effects of different training objectives and their interactions for network weight prediction tasks \citep{ashkenazi2022nern, soro2024diffusion, knyazev2023can}.
We demonstrate that better-performing models can be obtained through a weight reconstruction-only objective and these improvements can be compounded over multiple repetitions.
By iterating over the reconstruction process multiple times through repeatedly reconstructing the previously reconstructed weights, we establish an "inception"-like setup with nested improvements in each prediction round.
Furthermore, we discovered that using multi-objective loss functions during the reconstruction process can lead to contradictory training signals, resulting in limited performance gains.
Therefore, we propose a new training scheme that decouples the learning objectives into two phases: a reconstruction phase and a distillation phase, ensuring that each learning objective has the desired impact.
Remarkably, our approach results in
significant improvements compared to the selected state-of-the-art baseline
while improving compression efficiency.
Moreover, the proposed separation provides greater flexibility in the distillation step by allowing the use of powerful networks to be involved during the weight refinement.
All these improvements and insights led to several usage scenarios that were not possible or practical before, for example, obtaining a better-performing model through a predictor network that is much smaller than the original network, or iterative improving a given model through the reconstruction process.
The proposed approach provides a unique, even surprising, perspective for achieving model performance improvement that is different from existing weight manipulation approaches such as stochastic weight averaging \citep{guo2023stochastic, izmailov2018averaging}.
We also achieve storage compression via the predictor network which is orthogonal and composable with existing model pruning \citep{lee2019snip, liu2018rethinking, gao2021network, wang2021convolutional, he2023structured} and knowledge distillation approaches \citep{chen2020online, gou2021knowledge, chen2017learning, beyer2022knowledge}.

\section{Predicting Model Weights using Neural Representations}

There is a growing interest in predicting the weights of pre-trained models not only to enhance the memory efficiency of model storage but also to improve the throughput of model inference.
Existing solutions range from building high-fidelity auxiliary models~\citep{knyazev2023can} to learning parameter-efficient approximators~\citep{guo2023learning}.
However, in this paper, we focus on neural representations learning based on their flexibility and potential for producing effective, yet parameter-efficient, approximations for deep networks. 
The NeRN framework~\citep{ashkenazi2022nern}, introduced by Ashkenazi~\textit{et al.} first explored the idea of training INRs
and demonstrated its utility in compressing CNNs.
At its core, NeRN uses a 5-layer multilayer perceptron (MLP) $G_{\phi}$ with fixed hidden layer size to learn a mapping from an input tuple $($layer $\ell,$ filter $f,$ channel $c)$ to the corresponding $k\times k$ kernel in the original CNN model $F_{\theta}$. 
Note that the output size of the MLP also remains fixed at the largest kernel size and smaller kernels in the CNN are sampled from the middle
while
fully-connected or normalization layers are excluded based on their comparatively negligible parameter size.
By utilizing positional encodings, e.g., SIREN \citep{sitzmann2020implicit}, the input coordinates are projected into a higher dimensional space, allowing NeRN to learn higher frequency functions and improve the approximation capabilities of the predictor. 
Since there is no inherent smoothness in the ordering of filters in a CNN, permutation strategies are introduced to rearrange filters in the original model based on similarity, ensuring stable training of NeRN.

The central component of NeRN training is the reconstruction loss aimed at quantifying the disparity between the original network weights and those recovered using the predictor.
Regardless of the capacity of $G_{\phi}$, one might expect that the reconstruction loss should reliably converge to a meaningful approximation.
However, in practice, it has been found that NeRN is prone to training instabilities when the auxiliary model is smaller than the original network.
To mitigate these issues, the authors introduced additional loss terms inspired by existing knowledge distillation methods.
However, it is important to note that while the reconstruction loss does not require access to training data, 
the distillation loss does, making it impractical for scenarios where access to the training data is not available.
The overall objective function for NeRN can be expressed as
\begin{gather}
\mathcal{L}_{objective}=\mathcal{L}_{recon}+\alpha\mathcal{L}_{KD}+\beta\mathcal{L}_{FMD}, \text{with} \numberthis \\
\nonumber \mathcal{L}_{recon}=\frac{1}{|\mathbf{W}|}\|\mathbf{W}-\mathbf{\hat{W}}\|_{2}, \phantom{a}\mathcal{L}_{FMD}=\frac{1}{|\mathcal{B}|}\sum_{i\in\mathcal{B}}\sum_{l}\|\mathbf{a}^{l}_{i}-\mathbf{\hat{a}}^{l}_{i}\|_{2}, \phantom{a} \mathcal{L}_{KD}=\frac{1}{|\mathcal{B}|}\sum_{i\in\mathcal{B}}KL(\mathbf{a}^{\text{out}}_{i}, \mathbf{\hat{a}}^{\text{out}}_{i}),
\label{eq:nern_objective}
\end{gather}
where $\mathbf{W}=[\mathbf{w}^{0}, \mathbf{w}^{1}, ..., \mathbf{w}^{L}]$, 
represents a list of convolutional weight vectors for layer $\ell$ in the original network, and $\hat{\mathbf{W}}$ denotes the corresponding weights of the reconstructed weights.
The terms $\mathbf{a}^{\ell}_{i}$ and $\hat{\mathbf{a}}^{\ell}_{i}$ denote the ${L}_{2}$ normalized feature maps generated from the $i$-th sample in the minibatch $\mathcal{B}$ at layer $\ell$ for the original and reconstructed networks respectively.
Additionally, the logit distillation loss $\mathcal{L}_{KD}$ employs the Kullback-Leibler divergence to compare the output logits $\mathbf{a}^{\text{out}}_{i}$ and $\hat{\mathbf{a}}^{\text{out}}_{i}$ from the original and reconstructed networks for each sample $i$ in the minibatch $\mathcal{B}$.

\section{Proposed Approach}
\label{sec:main_observation}

In this section, we take a closer look at learning neural representations for pre-trained neural networks. 
Specifically, we explore the role of different training objectives in realizing an effective parameterization.
Based on our findings, we propose a new training scheme that circumvents the undesirable trade-off between accuracy and model compression 
rate, and simultaneously improve on both aspects. 

\subsection{Is the Reconstruction Objective All You Need?}
\label{sec:NeRN_Inception_Training}

A well-known limitation of existing approaches used for weight prediction is that they trade off accuracy to achieve parameter efficiency. 
This non-trivial compromise in model performance can be a critical bottleneck for practical applications.
While existing approaches attempt to recover the lost performance through the use of additional objectives, e.g., distillation as in NeRN, they lead to increased reconstruction errors, albeit providing improvements in the accuracy.
As this seems counter-intuitive, it naturally raises the question:
what is the relation between reconstruction error and expected model performance?

If we assume that reconstruction error, e.g., mean-squared error (MSE), is indeed an indication for performance, a straightforward strategy to improve performance would be to increase the capacity of the predictor network, allowing it to overfit to the original model weights.
To this end,
\begin{wrapfigure}{r}{0.55\textwidth}
    \vspace{-3mm}
	\centering
	\includegraphics[width=0.55\textwidth]{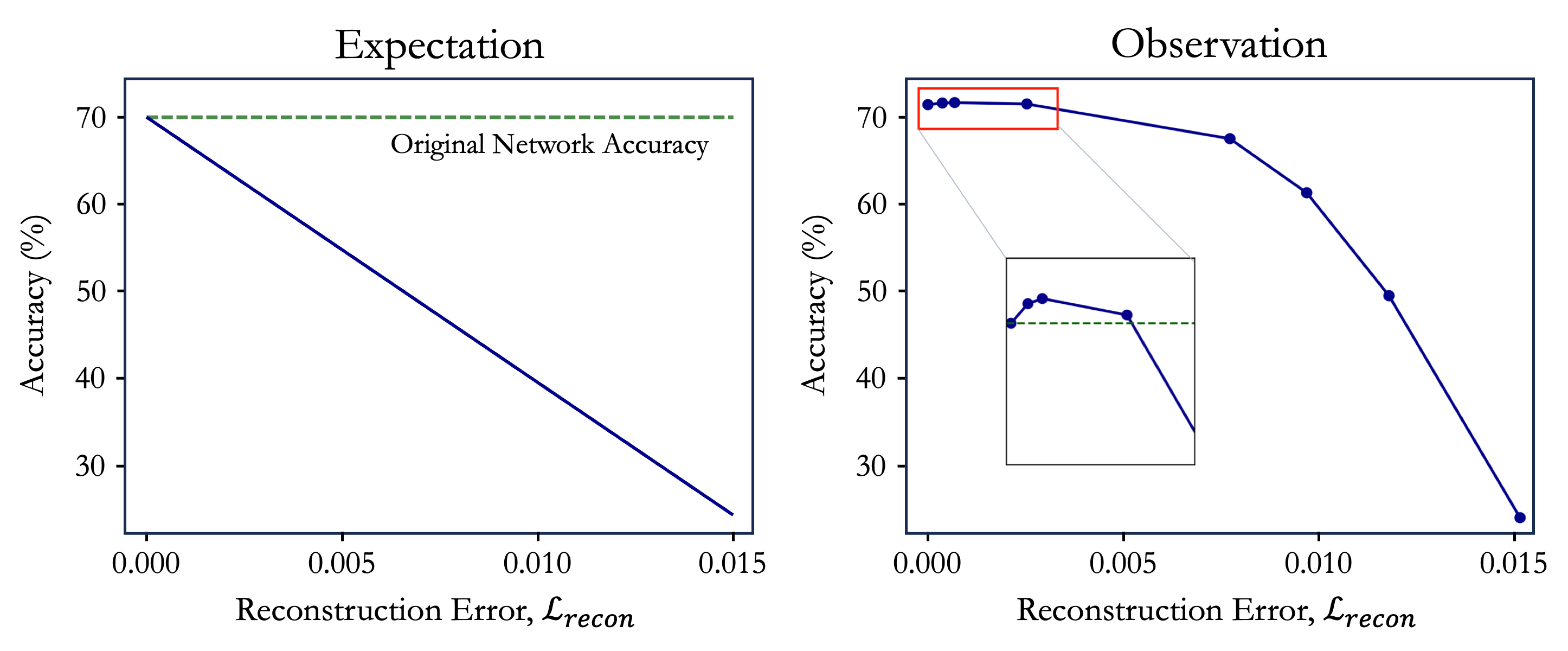}
     \vspace{-5mm}
	\caption{While one expects that the reconstruction error must approach zero to recover the true performance (left), we are able to find networks with non-zero (yet low) errors that not only match the true performance but even surpass it (right).}
    \label{fig:recon_illustration}
\vspace{-2mm}
\end{wrapfigure} we first empirically analyze how well we can recover the original network's performance, as we continually reduce the reconstruction error by increasing the predictor network capacity.
Note, in this analysis, we are not concerned about parameter-efficiency, and the neural representations are trained solely based on reconstruction error.
As shown in Figure~\ref{fig:recon_illustration} (left), one might expect that the parameterization should recover the performance of the original network as the reconstruction error approaches zero.
However, by using only the reconstruction loss and increasing the predictor network capacity, 
we empirically find weight parameterizations with
non-zero reconstruction errors that not only recover the true performance, but even surpass it (Figure~\ref{fig:recon_illustration} (right)).

\textit{How does reconstruction-only objective lead to better networks?} A well-known property of the MSE\begin{wrapfigure}{l}{0.35\textwidth}
    \vspace{-2mm}
	\centering
	\includegraphics[width=0.35\textwidth]{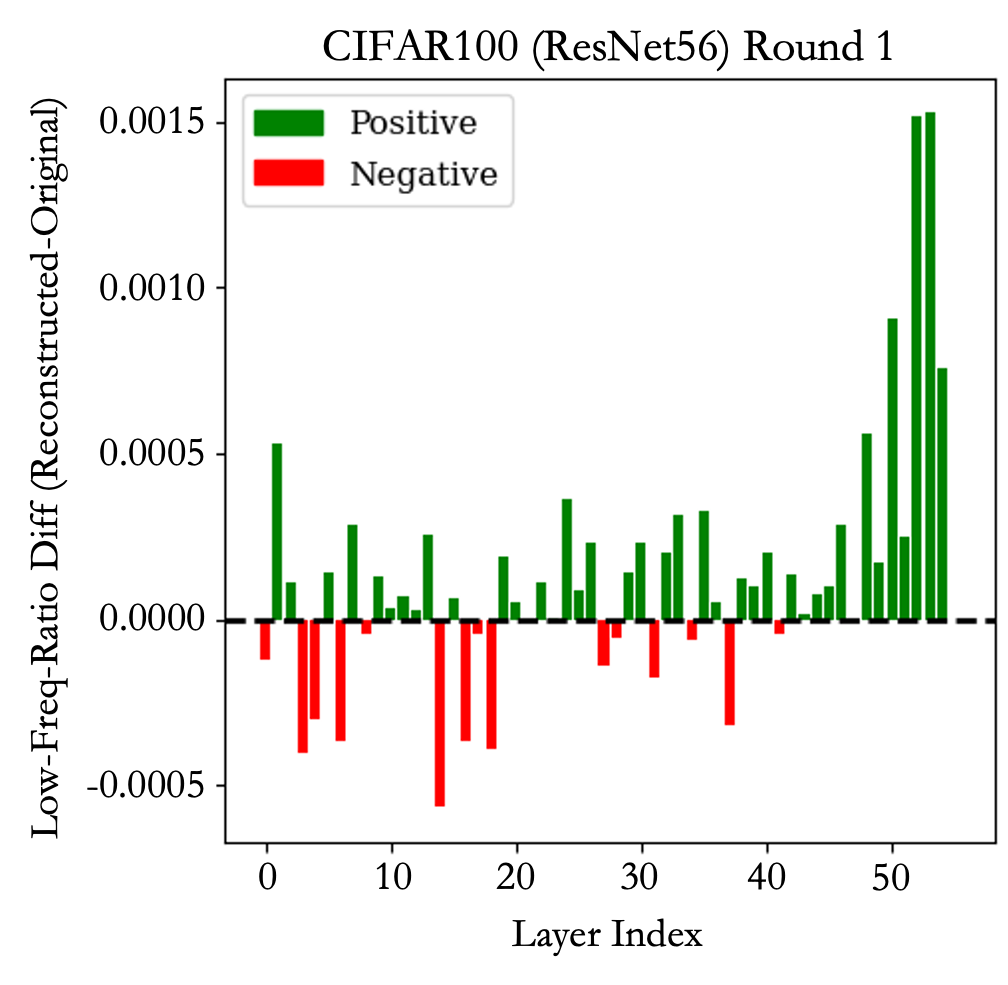}
     \vspace{-5mm}
	\caption{Measuring the dominance of low-frequency components in original and recovered networks.}
    \label{fig:Sratio}
\vspace{-4mm}
\end{wrapfigure} loss is that it tends to have a smoothing effect on the reconstructed weights~\citep{domingos2012few}.
To quantify this, we compare the weights of the reconstructed network (based on neural representations learned using only the reconstruction objective) with those of the original network, in terms of changes in the ratio of low-frequency components of their weight matrices measured as follows:
Let \(\mathbf{M}\) be the weight matrix of a given layer, shaped as \(\mathbf{M} \in \mathbb{R}^{n \times m}\).
Using singular value decomposition (SVD) on this matrix, we obtain $\mathbf{M} = \mathbf{U} \mathbf{\Sigma} \mathbf{V}^T$ where \(\mathbf{\Sigma} \in \mathbb{R}^{n \times m}\) is a diagonal matrix with singular values $\mathbf{s} = \left[\sigma_1, \sigma_2, \ldots, \sigma_k\right]$ on the diagonal and $k=\min(n, m)$.
The $S_{ratio}$ is then calculated as
\(S_{ratio} = \sum_{i=1}^{\lfloor k/2 \rfloor} \sigma_i^2 / \sum_{i=1}^k \sigma_i^2\).
This ratio measures the proportion of the total variance (energy) of the matrix \(\mathbf{M}\) that is captured by the first half of the singular values. A higher ratio indicates that the lower frequency components are more dominant. As illustrated in Figure \ref{fig:Sratio}, we show a per-layer difference of $S_{ratio}$ between the reconstructed and original weights. 

Upon close inspection, a clear trend emerges, where we find that the reconstructed weights correspond to more dominant low-frequency components compared to the original, particularly in the later layers, i.e., this smoothing leads to implicit rank reduction.
Interestingly, this observation corroborates with the recent finding that rank reduction can help improve the behavior of even large scale models \citep{sharma2023truth}.

Building upon our hypothesis about the effect of smoothing induced by the reconstruction objective, we ask the question:
\textit{can multiple rounds of progressive, neural representation learning further improve the performance of the reconstructed network?} To this end, we extend our previous experiment by training multiple generations of network parameterizations, wherein the first round predictor network attempts to reconstruct the original network, the second round predictor attempts to recover the reconstructed network from round 1, and so on.
\begin{wrapfigure}{r}{0.35\textwidth}
    \vspace{-5mm}
	\centering
	\includegraphics[width=0.35\textwidth]{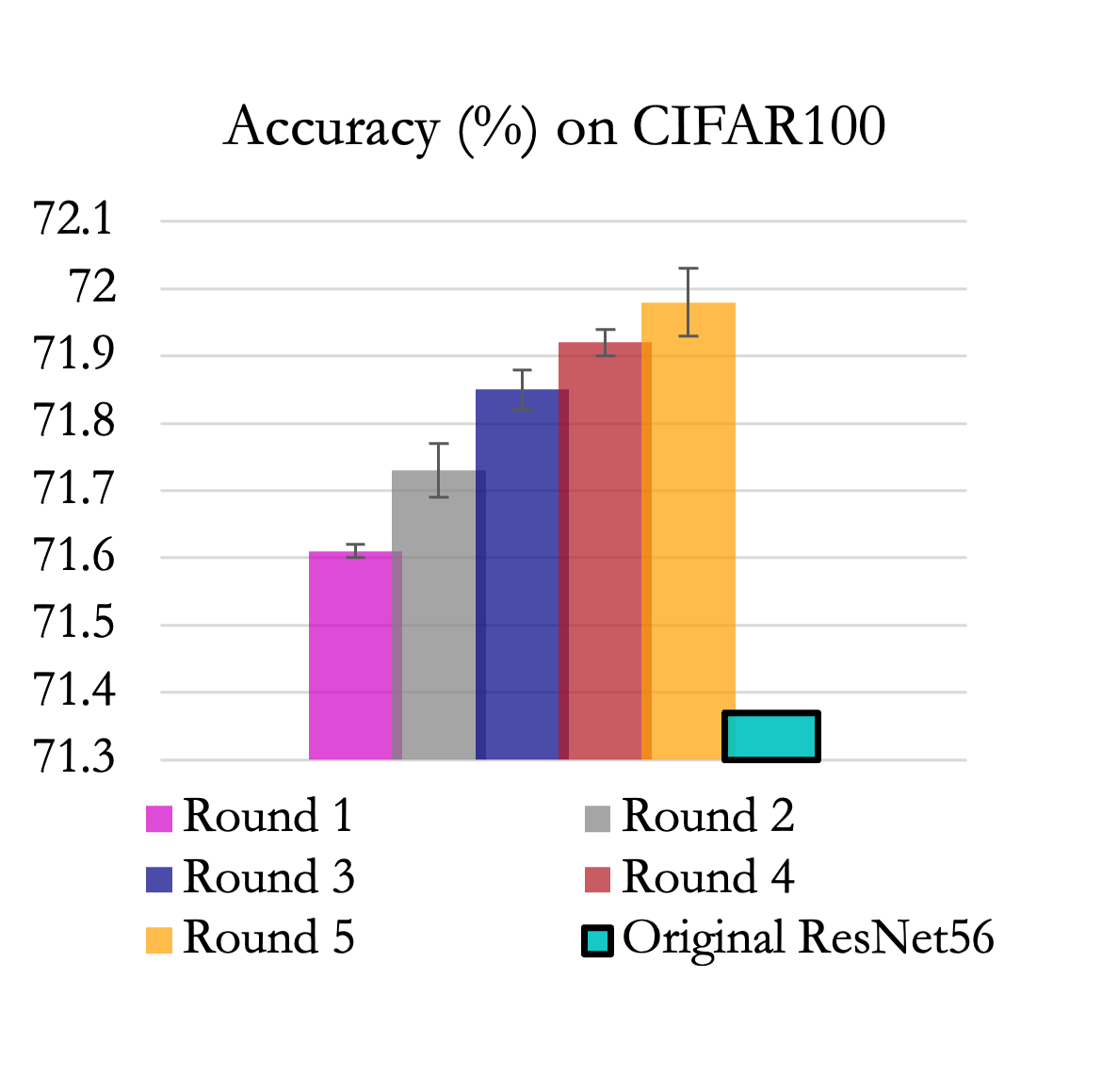}
     \vspace{-10mm}
	\caption{Evaluation of the reconstruction performance for each round of reconstruction.}
    \label{fig:NeRN_Inception}
\vspace{-2mm}
\end{wrapfigure} Such a process resembles an "inception"-like setup where the new predictor is built on top of the previously reconstructed weights. We find that, this simple procedure leads to further improvements over the original network's performance, albeit producing relatively higher reconstruction errors due to smoothing. As shown in Figure \ref{fig:NeRN_Inception}, this progressive training produces monotonic improvements in the accuracy, and in all cases superior to the original network's performance. 
While this observation holds for all architectures and datasets, once weights are sufficiently smoothed, we do not witness further improvements by using additional rounds of training (performance does not drop either). 
Given the additional complexity in training multiple neural representations, in practice, even $1$ or $2$ rounds of progressive training is beneficial.

\subsection{Distillation Improves Compression, But Only with Loss Decoupling}\label{sec:NeRN_Decoupled_Training}

So far, we inspected the behavior of the reconstruction loss, and demonstrated its surprising efficacy in enhancing model performance. Despite the observed performance improvement, we did not take parameter-efficiency into account for our analysis. However, in practice, an important motivation for using weight prediction networks is to achieve signification reduction in memory requirements for model storage, while not trading-off the performance unreasonably. While neural representations were originally leveraged with such a compression objective~\citep{ashkenazi2022nern}, their accuracy trade-off makes them a less preferred choice over other model compression (or reduction) strategies in practice. In this section, we show how the compression capability of neural representations can be enhanced. To this end, we take a closer look at the widely adopted distillation objective and how it interacts with the reconstruction loss. By doing so, we address the critical need to strike a balance between model complexity and efficiency, paving the way for more practical and resource-efficient neural networks.

\begin{figure}[htbp]
    \begin{subfigure}[b]{0.3\textwidth}
        \includegraphics[width=\textwidth]{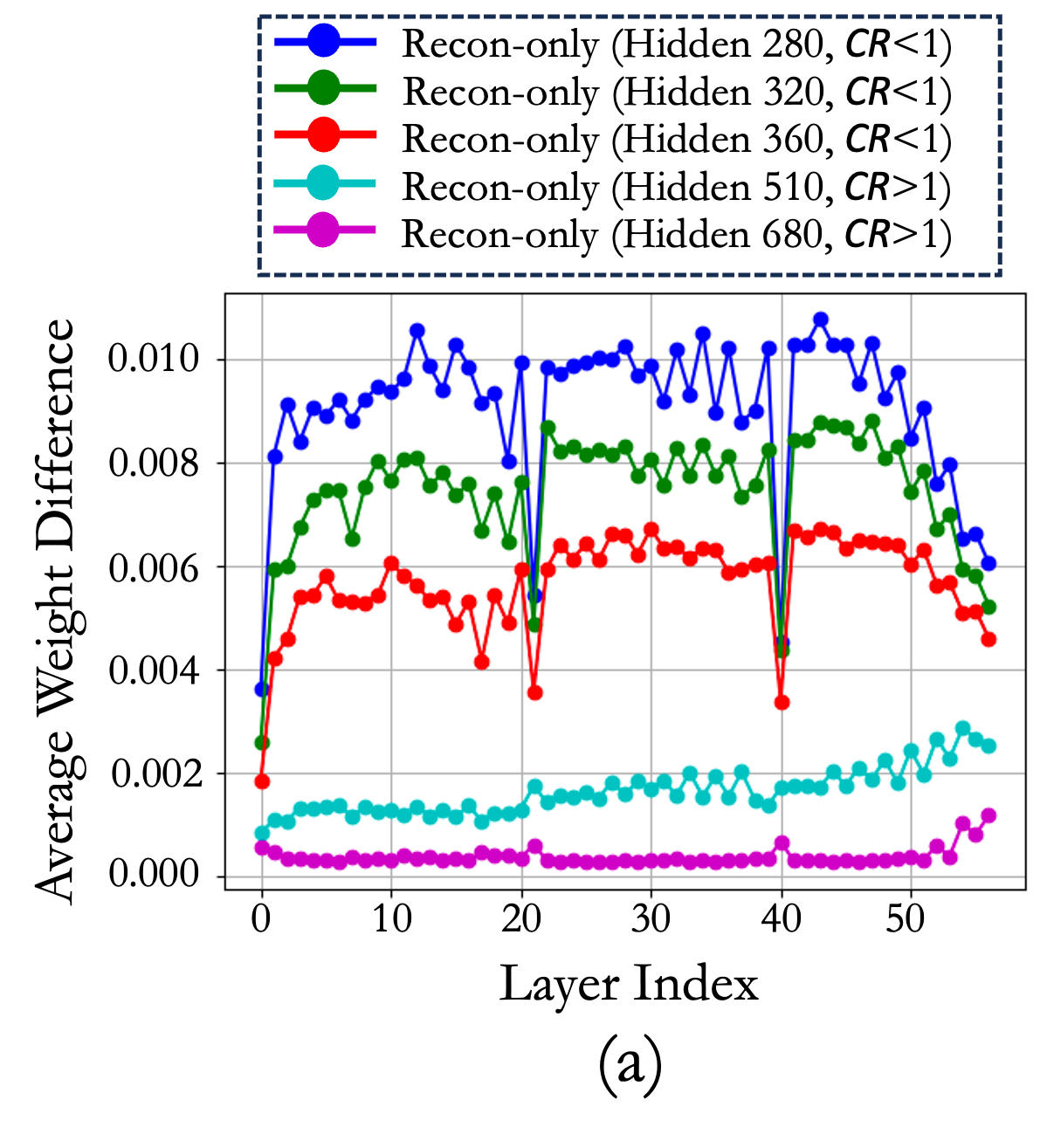}
    \end{subfigure}
    \begin{subfigure}[b]{0.3\textwidth}
        \includegraphics[width=\textwidth]{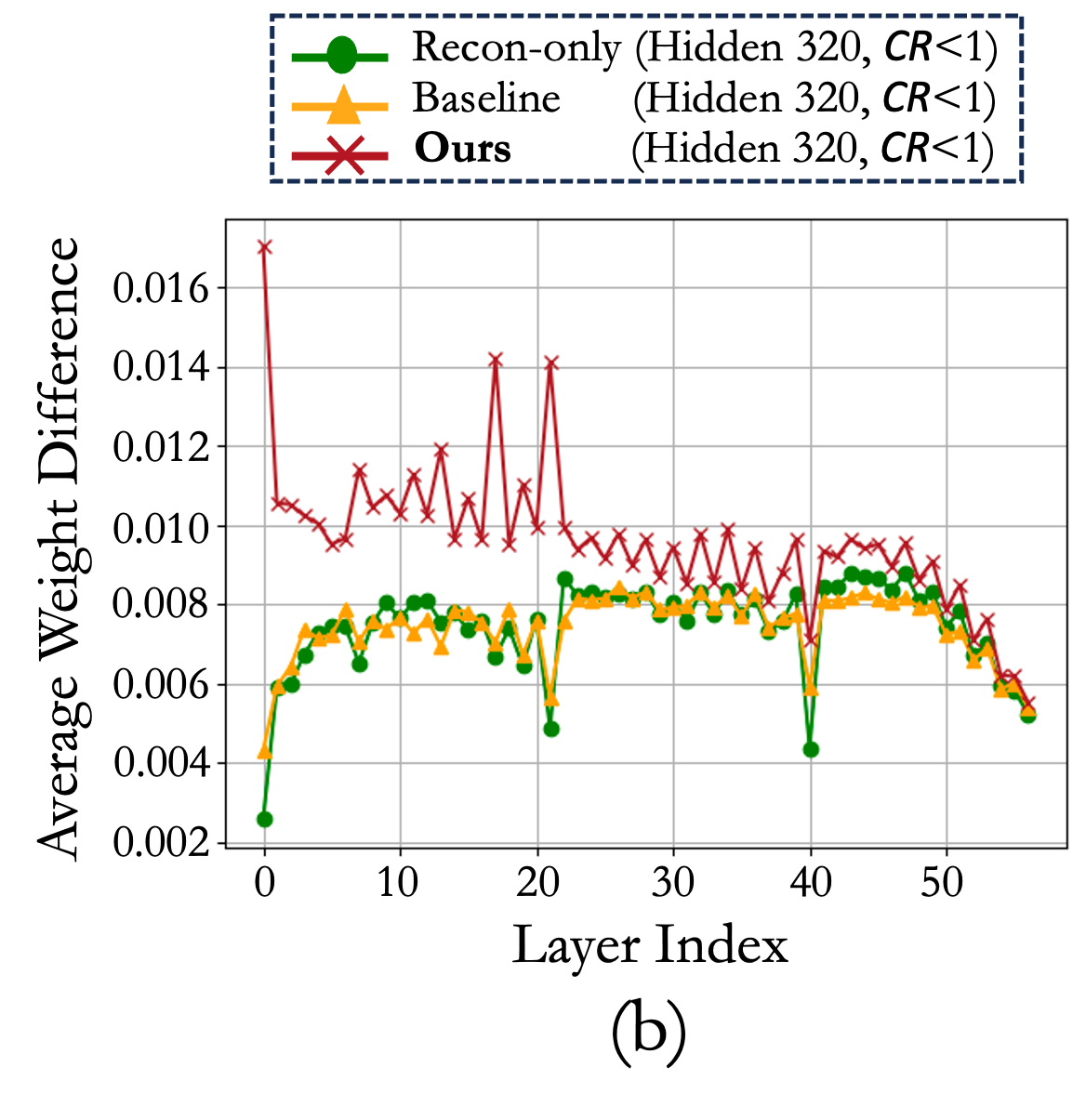}
    \end{subfigure}
    \begin{subfigure}[b]{0.4\textwidth}
        \includegraphics[width=\textwidth]{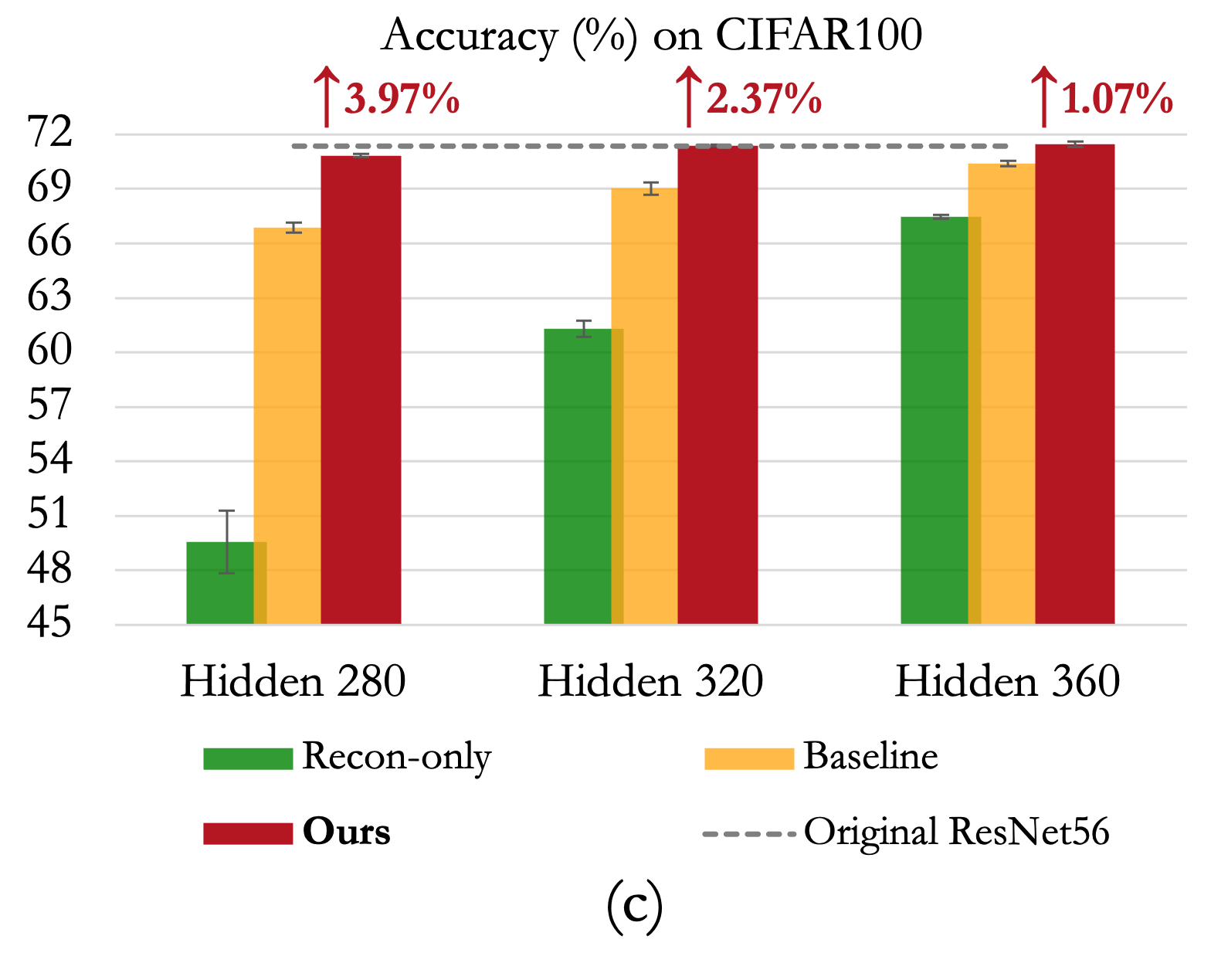}
    \end{subfigure}    
    \vspace{-6mm}
    \caption{(a) Average weight difference between predicted weights by Recon-only and original weights with varying hidden sizes. (b) Comparison among Recon-only, baseline, and ours. (c) Evaluation of the reconstruction performance for each method. $\uparrow$ represents our performance improvement over the baseline.}\label{NeRN_Decoupling}
\end{figure}

To begin with, we analyze the layer-wise weight differences between the reconstructed and original networks at varying predictor network sizes (Figure \ref{NeRN_Decoupling}(a)). Note, we use the term  ``\textsl{CR}'' to denote the \emph{compression ratio}, computed as the ratio of the learnable size of the predictor $S(P)$ to that of the original network $S(O)$. For instance, if the predictor network has $Q$ learnable parameters in order to recover an original network with $P$ parameters, then $\textsl{CR}\times100 = (Q/P)\times100\%$. As expected, smaller-sized predictor networks (i.e., higher compression) exhibit relatively large gaps and this can be attributed to the insufficient representation power. On the other hand, increasing the capacity of the predictor network leads to improved reconstruction performance, as depicted by the green bar in Figure \ref{NeRN_Decoupling}(c). However, insights from Section \ref{sec:NeRN_Inception_Training} indicate that, unless we increase the predictor capacity even further and perform progressive training, the reconstruction-only training cannot match the true performance. Hence, to recover the lost performance while also enabling parameter-reduction (i.e., CR < 1), the baseline NeRN approach~\citep{ashkenazi2022nern} incorporates an additional distillation objective during training. As illustrated by the orange bar in Figure \ref{NeRN_Decoupling}(c), the predictor network's performance can be significantly improved with the guidance of this distillation process. This can be attributed to the fact that the arbitrary perturbations in the reconstructed network induced by the reconstruction objective can make non-trivial changes to the decision rules, thus impacting the generalization performance of the network. Hence, additional guidance in terms of the prediction probabilities provides valuable task-relevant information.

While the baseline approach has proven effective, we observe that the role of the distillation objective is to primarily supplement the reconstruction loss.
However, with limited predictor network capacity, it is not possible to accurately recover the network weights. Hence, we argue that, \textit{compromising on the reconstruction fidelity can be acceptable, as long as the decision rules from the original network are preserved in the reconstructed network (i.e., emulate the predictions of the original model akin to traditional knowledge distillation)}. For instance, Figure \ref{NeRN_Decoupling}(b) illustrates the layer-wise discrepancies between the Recon-only and the baseline weights, in comparison to the original networks for the case of \textit{Hidden = 320} (number of neurons in each layer of the MLP for neural representations). It is apparent that the training process appears to be predominantly driven by the reconstruction loss, thus limiting its ability at high compression factors. While one can further emphasize the distillation term in \eqref{eq:nern_objective} by increasing $\alpha$, we find that it leads to training instabilities and the resulting network is far inferior.
This highlights the complimentary nature of the two objectives, but also how it can be challenging to combine them in practice.

To address this critical limitation, we propose to employ the two objectives in distinct training stages. Initially, we train the predictor network solely based on the reconstruction objective (Recon-only, $\mathcal{L}_{recon}$) and optionally perform progressive training when the predictor network capacity is high. In the second phase, we fine-tune the predictor stage 1 using only the distillation objective, $\mathcal{L}_{KD}$. As shown in Figure \ref{NeRN_Decoupling}(b), this allows the predictor to explore solutions that can in principle be different from the original network, but adhere to similar decision boundaries. Interestingly, we find that this two-stage optimization leads to significant differences in the early layers of the network, while still matching the later layers. This is intuitive since it is well known that the decision rules typically emerge in the later layers of a deep network. As we will demonstrate later, this decoupling of the training objectives not only demonstrates greater resilience to variations in the predictor network size, but also recovers or even surpasses the performance of the original network with predictors that are $>40\%$ smaller (see results in Table \ref{test_accuracy_nern+decoupling}).

\subsection{Improving Compression-Performance Trade-off via High-Performing Teachers}
\label{sec:dual_objective_through_high_teacher}

In section \ref{sec:NeRN_Inception_Training}, and \ref{sec:NeRN_Decoupled_Training}, we considered the performance or the compression objective independently. However, a natural next question is whether one can improve the compression vs. performance trade-off, and obtain additional improvements in both aspects. The proposed decoupled training offers flexibility in the second phase after the initial reconstruction objective is accomplished. One particular benefit of such a decoupling is that it enables the use of other high-performing models for guiding the distillation phase. 
We argue that, \textit{by leveraging guidance from a high-performing teacher, one can further improve the efficacy of decoupled training, thereby improving on the performance-compression trade-off.} In other words, through the proposed strategies, one can achieve non-trivial improvements to model accuracy for a fixed predictor network capacity, or easily push past the original network's performance via progressive reconstruction.

\begin{figure}[htb!]
	\centering
	\includegraphics[width=0.8\linewidth]{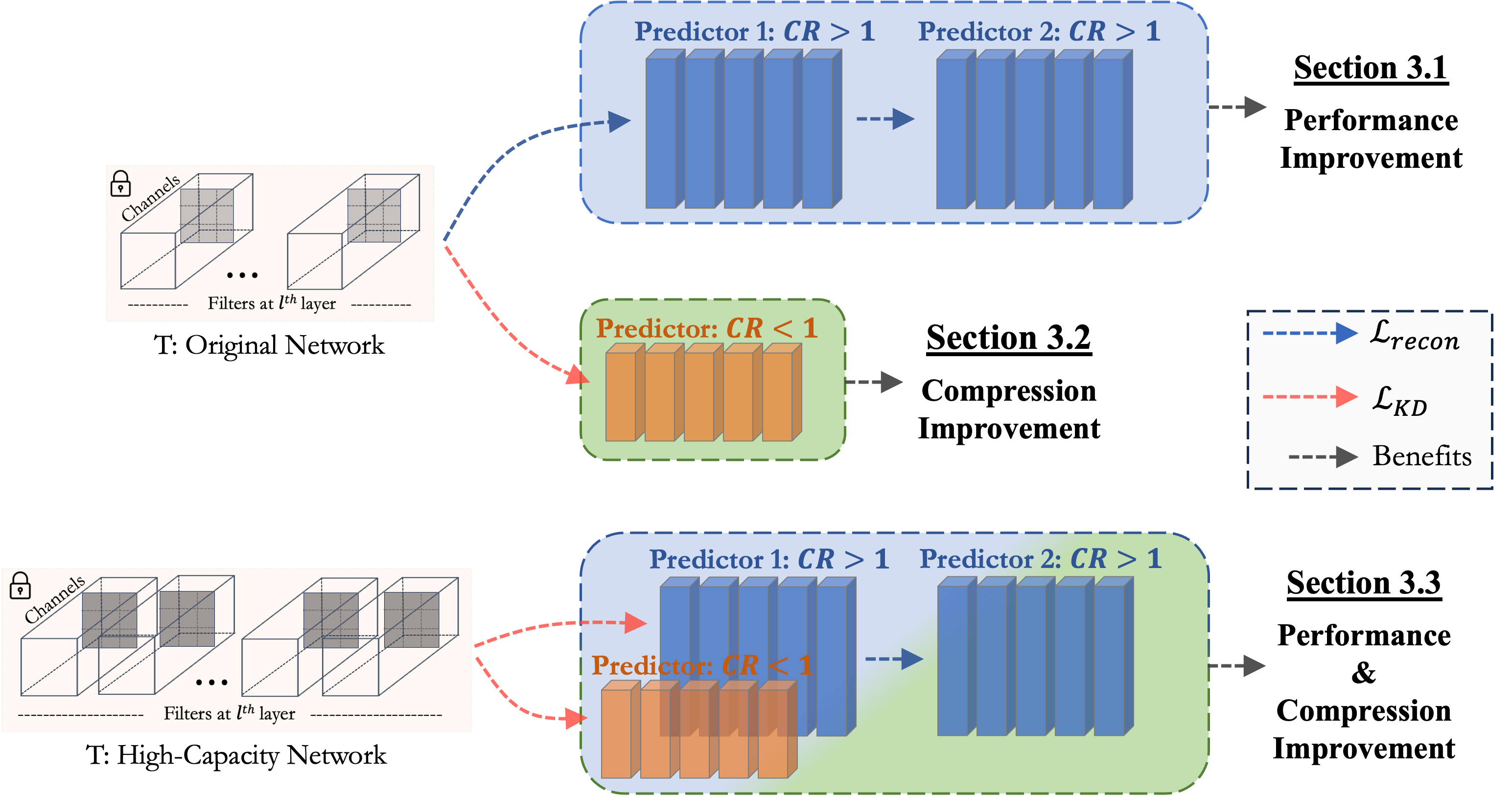}
	\caption{A summary of the explored training schemes and their benefits.
 We investigate the impact of reconstruction-only setup in Section \ref{sec:NeRN_Inception_Training}, and introduce the decoupled training with separate reconstructed and distillation stages in Section \ref{sec:NeRN_Decoupled_Training}, which improve storage compression via the more effective predictor networks. Lastly, we demonstrate that a high-capacity teacher can further facilitate both the compression and accuracy goals in Section \ref{sec:dual_objective_through_high_teacher}.}
 \label{fig:summary_of_works}
 \label{fig:contributions}
\end{figure}

This flexibility of our proposed approach goes beyond the decoupled training, as every component we have introduced can be combined with each other or with other methods (e.g., model pruning and standard knowledge distillation). For example, we can apply multiple rounds of reconstruction training to boost the starting point for the second distillation phase. Or the original model can be from a previously pruned model, where the proposed method can further enhance its performance.
The limitation is mostly computational as every additional step would incur additional training. To help put everything together, we provide a summary of our findings and components of the proposed approach in Figure \ref{fig:contributions}. We also illustrate their practical benefits, i.e., performance improvement and/or compression. In Section \ref{sec:Experiment}, we will provide empirical evidence to support each of the illustrated setups, namely, reconstruction-only training for large predictor networks, decoupled training with reconstruction and distillation objectives for parameter-efficient predictor networks, and finally decoupled training with high-capacity teachers at all predictor network capacities.

\section{Experiments}
\label{sec:Experiment} 

In this section, we present experiment results to support the observations made in Section \ref{sec:main_observation} and highlight the practical usage scenarios made possible with our proposed approach.
We use benchmark datasets, including CIFAR-10, CIFAR-100 \citep{krizhevsky2009learning}, STL-10 \citep{coates2011analysis}, and ImageNet \citep{deng2009imagenet}.
Following a similar setup outlined in \citep{ashkenazi2022nern}, we also focus on ResNet architectures \citep{he2016deep} as the reconstructed models.
Beyond evaluating the in-distribution performance of the reconstructed models, we also assess their robustness using out-of-distribution (OOD) datasets such as CIFAR10-C, CIFAR100-C, and ImageNet-R, 
and two popular adversarial attacks, FGSM and I-FGSM \citep{goodfellow2014explaining}.
Detailed information about the datasets and training setups is provided in the Appendix \ref{detailed_info_training}.

\subsection{Enhancing Performance Through Progressive Reconstruction - \texorpdfstring{\textsl{CR}$>$1}{CR > 1}}\label{sec:inception_training}

Firstly, we discovered that the reconstructed model using only the reconstruction loss performs better in testing when weights are predicted by a large predictor ($\textsl{CR}>1$).
The reconstruction loss alone enabled the model to improve its performance slightly, with gains ranging from 0.1\% to 0.3\%.
Interestingly, additional rounds of weight prediction further enhanced performance beyond the initial reconstructed model, ultimately achieving gains of up to 0.6\%. Table \ref{results_inception_nern} presents the results of the reconstructed models at each round.
For example, the first round (\textit{Round 1}) aims to reconstruct the original network.
Then, the second round (\textit{Round 2}) predicts the weights of the first-round reconstructed model.
This process continues progressively through multiple generations until performance improvements plateau.
As each round progresses, the predicted weights deviate more from the original solution, as indicated by the increased reconstruction loss. 
Additionally, the solutions found in each round generally exhibit greater robustness against adversarial attacks than the original network.

\begin{table*}[htb!]
\centering
\caption{Evaluation performance of large predictor networks via progressive reconstruction. The reported results are the mean values over three runs.}
\label{results_inception_nern}
\vspace{0.2em}
\centering 
\scalebox{0.85}{
\begin{tabular}{lcccccc}  
\cmidrule[1.5pt](lr){1-7}
\multirow{2}{*}{\makecell{CIFAR10\vspace{5pt}}} & \multirow{2}{*}{\makecell{Original\\ResNet20}} & \multicolumn{5}{c}{Hidden 300 (\textsl{CR}$>$1)} \\ 
\cmidrule(lr){3-7} & & \multicolumn{5}{c}{\textit{Round 1} \quad \textit{Round 2} \quad \textit{Round 3} \quad \textit{Round 4}}  \\
\midrule
Accuracy ($\uparrow$, \%) & 91.69\%  & \multicolumn{5}{c}{\textbf{91.79} \quad\quad \textbf{91.88} \quad\quad \textbf{91.98} \quad\quad \textbf{92.00}} \\
$\mathcal{L}_{recon}$ & - & \multicolumn{5}{c}{0.00332 \quad 0.00414 \quad 0.00500 \quad 0.00547}   \\
OOD ($\uparrow$, \%) & \textbf{70.49} & \multicolumn{5}{c}{69.45 \quad\quad 69.32 \quad\quad 68.82 \quad\quad 68.85} \\
FGSM ($\uparrow$, \%) & 76.41 & \multicolumn{5}{c}{\textbf{77.16} \quad\quad 77.03 \quad\quad 77.08 \quad\quad 77.08} \\
I-FGSM ($\uparrow$, \%) & 75.02 & \multicolumn{5}{c}{75.56 \quad\quad 75.49 \quad\quad 75.59 \quad\quad \textbf{75.69}} \\
\cmidrule[1.5pt](lr){1-7}
\multirow{2}{*}{\makecell{STL10\vspace{5pt}}} & \multirow{2}{*}{\makecell{Original\\ResNet56}} & \multicolumn{5}{c}{Hidden 680 (\textsl{CR}$>$1)} \\ 
\cmidrule(lr){3-7} & & \multicolumn{5}{c}{\textit{Round 1} \quad \textit{Round 2} \quad \textit{Round 3} \quad  \textit{Round 4}} \\
\midrule
Accuracy ($\uparrow$, \%) & 76.31\% & \multicolumn{5}{c}{\textbf{76.40} \quad\quad \textbf{76.44} \quad\quad \textbf{76.51} \quad\quad \textbf{76.54}} \\
$\mathcal{L}_{recon}$ & - & \multicolumn{5}{c}{0.00120 \quad 0.00117 \quad 0.00125 \quad 0.00128}   \\
FGSM ($\uparrow$, \% & 39.28 & \multicolumn{5}{c}{\textbf{39.36} \quad\quad 39.27 \quad\quad 39.30 \quad\quad 39.32} \\
I-FGSM ($\uparrow$, \%) & 36.12 & \multicolumn{5}{c}{36.13 \quad\quad 36.24 \quad\quad \textbf{36.26} \quad\quad 36.17} \\ 
\cmidrule[1.5pt](lr){1-7}
\multirow{2}{*}{\makecell{CIFAR100\vspace{5pt}}} & \multirow{2}{*}{\makecell{Original\\ResNet56}} & \multicolumn{5}{c}{Hidden 680 (\textsl{CR}$>$1)} \\ 
\cmidrule(lr){3-7} & & \textit{Round 1} & \textit{Round 2} & \textit{Round 3} & \textit{Round 4} & \textit{Round 5}\\
\midrule
Accuracy ($\uparrow$, \%) & 71.37  & \textbf{71.61} & \textbf{71.73} & \textbf{71.85} & \textbf{71.92} & \textbf{71.98} \\
$\mathcal{L}_{recon}$ & - & 0.00068 & 0.00082 & 0.00088 & 0.00095 & 0.0010 \\
OOD ($\uparrow$, \%) & 44.70 & 44.47 & \textbf{44.75} & 44.50 & 44.29 & 44.47 \\
FGSM ($\uparrow$, \%) & 44.69 & 44.83 & 44.80 & 45.02 & \textbf{45.23} & 45.18 \\
I-FGSM ($\uparrow$, \%) & 39.31 & 40.91 & 40.82 & 41.04 & \textbf{41.21} & 41.04 \\
\bottomrule
\end{tabular}}
\end{table*}

\subsection{Achieving Greater Compression with Decoupled Training - \texorpdfstring{\textsl{CR}$<$1}{CR < 1}}\label{sec:decouple_training}

Shifting our focus from the previous section, where we used large predictors (\textsl{CR}>1) to improve model performance, we now aim to further compress the predictor size without significantly sacrificing performance.
The current training of predictors involves three objectives, as seen in Equation \ref{eq:nern_objective}, which guide the predictor to effectively mimic the original network.
However, in Figure \ref{NeRN_Decoupling}(b), we observed that the solution found by Equation \ref{eq:nern_objective} was only locally perturbed from the solution in Recon-only model, indicating a restrictive role of the distillation loss.
Therefore, we highlight the impact of a simple decoupling of the learning objectives into two phases: a reconstruction phase and a distillation phase, achieving substantial compression gains.
To this end, we fine-tune the predictor for a small number of iterations using only $\mathcal{L}_{KD}$, starting from the solution derived by the Recon-only models. 

We found that even starting from an inferior point (24.20\% from Recon-only on CIFAR-100 in Table \ref{test_accuracy_nern+decoupling}), performance can be significantly recovered to 69.31\% with only $\mathcal{L}_{KD}$ in the second phase, representing a 9\% improvement over the Baseline.
Furthermore, we achieve a compression ratio of approximately $57\%$ (Hidden $360$) while surpassing the original network's performance. 
With these significant improvements compared to the existing baseline, we can achieve better (CIFAR10, CIFAR100) or comparable performance (STL10, ImageNet) with a predictor network smaller than the original network. In the case of ImageNet, we can obtain a $15\%$ compression with only a $3\%$ performance drop compared to the 8\% for the baseline.

\begin{table*}[htb!]
\centering
\caption{Evaluation performance of small predictor networks via decoupled training. The reported results represent the mean values over three runs except for ImageNet experiments. Recon-only ($\mathcal{L}_{recon}$), Baseline ($\mathcal{L}_{recon}+\mathcal{L}_{KD}+\mathcal{L}_{FMD}$), and \textbf{Ours} ($\mathcal{L}_{KD}$ only in the second phase).}
\label{test_accuracy_nern+decoupling}
\vspace{1.0em}
\centering 
\scalebox{0.85}{
\begin{tabular}{lcccc}  

Method & \multicolumn{4}{c}{Recon-only / Baseline / \textbf{Ours}} \\ 
\cmidrule[1.5pt](lr){1-5}   
CIFAR10 & \makecell{Original\\ResNet20} & {\makecell{Hidden 120\\(\textsl{CR}$\times$100 $\approx$27\%)}} & {\makecell{Hidden 140\\(\textsl{CR}$\times$100 $\approx$35\%)}} &  {\makecell{Hidden 180\\(\textsl{CR}$\times$100 $\approx$53\%)}}\\
\midrule
Accuracy ($\uparrow$, \%) & 91.69 & 75.75 / 87.99 / \textbf{90.75} & 85.64 / 89.67 / \textbf{91.34} & 90.03 / 91.26 / \textbf{91.75} \\
OOD ($\uparrow$, \%) & 70.49 & 50.50 / 65.00 / \textbf{68.84} & 60.08 / 67.19 / \textbf{69.95} & 66.34 / 69.75 / \textbf{70.48} \\ 
FGSM ($\uparrow$, \%) & 76.41 & 59.76 / 72.73 / \textbf{75.38} & 70.26 / 74.96 / \textbf{75.83} & 74.78 / 76.01 / \textbf{76.27} \\
I-FGSM ($\uparrow$, \%) & 75.02 & 58.87 / 71.48 / \textbf{73.77} & 69.05 / 73.58 / \textbf{74.21} & 73.39 / 74.44 / \textbf{74.83} \\
\cmidrule[1.5pt](lr){1-5}
CIFAR100 & \makecell{Original\\ResNet56} & {\makecell{Hidden 220\\(\textsl{CR}$\times$100 $\approx$24\%)}} & {\makecell{Hidden 280\\(\textsl{CR}$\times$100 $\approx$36\%)}} &  {\makecell{Hidden 360\\(\textsl{CR}$\times$100 $\approx$57\%)}}\\
\midrule
Accuracy ($\uparrow$, \%) & 71.37 & 24.20 / 60.94 / \textbf{69.31} & 49.55 / 66.87 / \textbf{70.84} & 67.48 / 70.39 / \textbf{71.46}  \\
OOD ($\uparrow$, \%) & 44.70 & 13.01 / 38.59 / \textbf{43.76} & 27.08 / 42.00 / \textbf{44.61} & 40.21 / 44.21 / \textbf{45.14} \\ 
FGSM ($\uparrow$, \%) & 43.69 & 14.65 / 40.30 / \textbf{45.35} & 30.76 / 43.51 / \textbf{44.95} & 43.28 / 44.82 / \textbf{45.01} \\
I-FGSM ($\uparrow$, \%) & 39.31 & 14.14 / 38.73 / \textbf{42.61} & 29.38 / 41.41 / \textbf{41.95} & 40.74 / \textbf{41.78} / 41.12 \\
\cmidrule[1.5pt](lr){1-5}
STL10 & \makecell{Original\\ResNet56} & {\makecell{Hidden 280\\(\textsl{CR}$\times$100 $\approx$36\%)}} & {\makecell{Hidden 320\\(\textsl{CR}$\times$100 $\approx$46\%)}} &  {\makecell{Hidden 360\\(\textsl{CR}$\times$100 $\approx$57\%)}}\\
\midrule
Accuracy ($\uparrow$, \%) & 76.31 & 69.41 / 74.99 / \textbf{76.02}  & 73.36 / 75.64 / \textbf{76.25} & 74.81 / 75.74 / \textbf{76.26} \\
FGSM ($\uparrow$, \%) & 39.28 & 36.27 / 39.69 / \textbf{39.82} & 38.89 / \textbf{39.83} / 39.28 & 38.53 / \textbf{39.77} / 39.21  \\
I-FGSM ($\uparrow$, \%) & 36.12 & 33.60 / 36.33 / \textbf{36.61} & 35.44 / \textbf{36.27} / 35.96 & 35.30 / \textbf{36.40} / 35.96 \\
\cmidrule[1.5pt](lr){1-5}
ImageNet & \makecell{Original\\ResNet18} & {\makecell{Hidden 700\\(\textsl{CR}$\times$100$ \approx$15\%)}} & {\makecell{Hidden 1024\\(\textsl{CR}$\times$100$ \approx$31\%)}} & {\makecell{Hidden 1372\\(\textsl{CR}$\times$100 $\approx$55\%)}} \\
\midrule
Accuracy ($\uparrow$, \%) & 69.76 & 51.10 / 61.91 / \textbf{66.48} & 65.69 / 67.32 / \textbf{68.68} & 68.81 / 68.87 / \textbf{69.32} \\
OOD ($\uparrow$, \%) & 33.07 & 19.87 / 25.59 / \textbf{30.25} & 28.36 / 30.90 / \textbf{32.17} & 31.72 / 32.54 / \textbf{32.82}   \\ 
FGSM ($\uparrow$, \%) & 57.83 & 39.99 / 50.19 / \textbf{53.94} & 53.82 / 55.67 / \textbf{56.69} & 56.96 / 57.14 / \textbf{57.40}  \\
I-FGSM ($\uparrow$, \%) & 57.15 & 38.42 / 49.63 / \textbf{53.28} & 53.23 / 55.06 / \textbf{56.09} & 56.32 / 56.43 / \textbf{56.76} \\

\bottomrule
\end{tabular}}
\end{table*}

\subsection{Distillation-Driven Compression and Performance Enhancement - \texorpdfstring{$\textsl{CR}<1$ and $\textsl{CR}>1$}{CR < 1 and CR > 1}}

In Sections \ref{sec:inception_training} and \ref{sec:decouple_training}, we explored two distinct avenues:
improving model performance with large predictors ($\textsl{CR}>1$) and achieving greater compression with small predictors ($\textsl{CR}<1$), respectively. In this section, we seek to simultaneously improve on both of the objectives.
Leveraging the flexibility of our decoupled training approach, we can utilize the superior guidance provided by a high-capacity teacher network to enhance parameter efficiency and produce high-fidelity representations. To this end, we employ ResNet50 as a teacher network, which has a size of 90.43MB with $78.48\%$ accuracy on CIFAR100, and the reconstructed network (ResNet56) with a size of 3.25MB.
As the teacher network is used only during the training in the second phase, the computational overhead and larger parameter of the teacher do not affect either the predictor network or the reconstructed model. 

Firstly, we present the results of a parameter-efficient predictor guided by the teacher network in Table \ref{test_accuracy_with_high_teachers_small_net}. For a fair comparison, the Baseline model is also trained using $\mathcal{L}_{KD}$ under the same teacher's supervision, as $\mathcal{L}_{FMD}$ is applicable only to architectures that are identical between the student and the teacher. The results in the table support the evidence that a high-performing teacher network can improve the efficiency of the predictor. For instance, in the `Hidden 280' scenario, our method's performance `with guidance' reveals an accuracy of $72.06\%$. This not only surpasses `without guidance' but also outperforms the original network itself.


\begin{table*}[htb!]
\centering
\caption{Evaluation performance of \textbf{parameter-efficient predictor networks with guidance} from a high-performing teacher network (ResNet50). We compare the effect of including the distillation objective to both the baseline ($\mathcal{L}_{recon}+\mathcal{L}_{KD}$), and the proposed approaches ($\mathcal{L}_{KD}$ only in the second phase). In each case, we show the results for \textit{without guidance} / \textit{with guidance}.}
\label{test_accuracy_with_high_teachers_small_net}
\vspace{1.0em}
\centering 
\renewcommand{\arraystretch}{1.3}
\resizebox{0.99\textwidth}{!}{
\begin{tabular}{lccccccc}  
\cmidrule[1.5pt](lr){1-8} 
CIFAR100 & \makecell{Original\\ResNet56} & \multicolumn{2}{c}{\makecell{Hidden 220\\(\textsl{CR}$\times$100 $\approx$24\%)}} & \multicolumn{2}{c}{\makecell{Hidden 280\\(\textsl{CR}$\times$100 $\approx$36\%)}} &  \multicolumn{2}{c}{\makecell{Hidden 360\\(\textsl{CR}$\times$100 $\approx$57\%)}}\\
\cline{3-8}
&&Baseline&Ours&Baseline&Ours&Baseline&Ours \\
\midrule
Accuracy ($\uparrow$, \%) & 71.37 & 60.94 / 58.30 & 69.31 / \textbf{70.25} & 66.87 / 66.21 & 70.84 / \textbf{72.06} & 70.39 / 70.94 & 71.46 / \textbf{72.91} \\
OOD ($\uparrow$, \%) & 44.70 & 38.59 / 35.45 & 43.76 / \textbf{44.66} & 42.00 / 41.13 &44.61 / \textbf{46.20} & 44.21 / 44.37 & 45.14 / \textbf{47.12} \\ 
FGSM ($\uparrow$, \%) & 43.69 & 40.30 / 37.01 & 45.35 / \textbf{46.69} & 43.51 / 42.65 & 44.95 / \textbf{47.32} & 44.82 / 45.53 & 45.01 / \textbf{48.24} \\
I-FGSM ($\uparrow$, \%) & 39.31 & 38.73 / 35.68 & 42.61 / \textbf{44.29} & 41.41 / 40.66 & 41.95 / \textbf{44.29} & 41.78 / 42.61 & 41.12 / \textbf{44.59} \\
\bottomrule
\end{tabular}}
\end{table*}


Interestingly, we observe that increasing the predictor size, particularly when parameter efficiency is not a critical factor, can lead to significant performance gains. This is likely due to the ability of a larger predictor to capture higher-fidelity representations of the teacher model. As shown in Table \ref{test_accuracy_with_high_teachers_large_net}, our method achieves superior performance levels the Baseline method cannot reach. Notably, our best-performing model achieves an accuracy of \textbf{73.95\%}, outperforming the conventional KD approach, a student (ResNet56) is trained from scratch using the guidance of ResNet50 teacher network with the $\mathcal{L}_{KD}$ loss, achieving $73.60\%$.

Building on the observation, and aligning with the idea presented in Section \ref{sec:inception_training} that additional training complexity can further improve performance, we proceed with one round of progressive reconstruction targeting our best-performing model (\textbf{73.95\%}). This resulted in a further improvement to \textbf{74.15\%}.

\begin{table*}[htb!]
\centering
\renewcommand{\tabcolsep}{6pt} 
\caption{Evaluation performance of \textbf{large predictor networks with guidance} from a high-performing teacher network (ResNet50). We compare the effect of including the distillation objective on both the baseline ($\mathcal{L}_{recon}+\mathcal{L}_{KD}$) and the proposed approaches ($\mathcal{L}_{KD}$ only in the second phase). In each case, we show the results for \textit{without guidance} / \textit{Baseline with guidance} / \textit{Ours with guidance}.}
\label{test_accuracy_with_high_teachers_large_net}
\vspace{1.0em}
\centering 
\resizebox{0.99\textwidth}{!}{
\begin{tabular}{lcccc}  
Method & \multicolumn{4}{c}{Recon-only / Baseline / \textbf{Ours}} \\ 
\cmidrule[1.5pt](lr){1-5}   
CIFAR100 & \makecell{Original\\ResNet56} & {\makecell{Hidden 510\\(\textsl{CR}$>$1)}} & {\makecell{Hidden 680\\(\textsl{CR}$>$1)}} &  {\makecell{Hidden 750\\(\textsl{CR}$>$1)}}  \\
\midrule
Accuracy ($\uparrow$, \%) & 71.37 & 71.45 / 72.02 / \textbf{73.41} & 71.61 / 71.80 / \textbf{73.95} & 71.56 / 71.89 / \textbf{73.82}  \\
OOD ($\uparrow$, \%) & 44.70 & 44.33 / 45.18 / \textbf{47.62} & 44.47 / 45.16 / \textbf{47.61} & 44.93 / 44.66 / \textbf{47.74}  \\ 
FGSM ($\uparrow$, \%) & 43.69 & 44.32 / 44.02 / \textbf{48.75} & 44.83 / 44.10 / \textbf{49.19} &  44.58 / 44.74 / \textbf{49.22} \\
I-FGSM ($\uparrow$, \%) & 39.31 & 40.24 / 39.82 / \textbf{45.24} & 40.91 / 39.94 / \textbf{45.26} &  40.35 / 40.55 / \textbf{45.53}  \\
\bottomrule
\end{tabular}}
\end{table*}

\section{Related Works}
\label{sec:related_works}

\noindent\textbf{Weight generation} methods utilizing implicit neural representations (INR) \citep{ashkenazi2022nern}, transformer \citep{knyazev2023can}, and diffusion model \citep{soro2024diffusion} for predicting model weights.
NeRN \citep{ashkenazi2022nern} produces an accurate reconstruction of a single network, whereas some other approaches focus on representing the distribution of weights or part of the overall weights.

\noindent\textbf{Weight space manipulation} provides a direct way to alter model behavior and comes in a variety of flavors. 
The recent development of ever larger models \citep{shoeybi2019megatron} makes fine-tuning existing models increasingly challenging, as a result, post-training model merging \citep{matena2022merging, tam2023merging, ilharco2022editing} are becoming increasingly popular that combines existing available models for performance enhancement.
Besides simply merging the weight, other types of weight manipulation also yield interesting results, e.g., improved model reasoning capability selectively reducing layer rank \citep{sharma2023truth}.
Apart from manipulating pretrained weights, there are also significant efforts leveraging operation on weight during training, such as stochastic weight averaging \citep{izmailov2018averaging, guo2023stochastic}.

\noindent\textbf{Implicit neural representations} 
or INR \citep{sitzmann2020implicit, tancik2020fourier} are initially designed for representing low-dimensional data (e.g., 2D or 3D) with complex and potential high-frequency signals.
Recently, INR has then been utilized for a variety of domains, e.g., from uncovering correlation in scientific data \citep{chitturi2023capturing} to estimating human pose \citep{yen2021inerf}.
In this work, the predictor network utilizes INR for predicting filter weights for model reconstruction. 

\noindent\textbf{Knowledge distillation and pruning}
has been widely adopted for reducing model size while preserving model performance. Knowledge distillation \citep{chen2020online, gou2021knowledge, chen2017learning, beyer2022knowledge} utilizes a more capable teacher network to transfer prediction behavior to the student network.
The pruning techniques \citep{lee2019snip, liu2018rethinking, gao2021network, wang2021convolutional, he2023structured} aim to remove non-essential or potentially duplicated functionality in the network and reduce the overall parameter counts.


\section{Discussion and Future Work}
\label{sec:discussion}
In this work, we identified effective strategies that significantly improve the accuracy of the reconstructed model and compression ratio for predictor networks through exploring various trade-offs in the parameterization of model weights with neural representation.
While effective, one area of the limitations is that the current predictor only works with CNN architectures, restricting its usage. 
Moreover, despite the flexibility of the proposed protocols that can be combined or re-applied, the necessary additional steps incur more training runs which can lead to a significant increase in computation cost and complexity.
However, the increased effort may be worth it to support edge applications where the benefits are multiplied by the number of deployed instances.
Still, to help address these challenges, we need methods that can predict weights for diverse architectures and are ideally more efficient when the target model grows in size and complexity.
Another interesting direction that is worthy of further investigation is the relationship between the frequency component of the weights and the model's generalization ability.
Could we directly alter the original weights to achieve a similar effect without the need to train a predictor model?
Or can we potentially use that insight during training as a regularization that directly improves models' generalizability?

This work reveals a unique perspective for model improvement and compression through the lens of weight parameterization, which is orthogonal to (i.e., can be arbitrarily combined with) traditional model performance improvement strategies such as stochastic weight averaging or model compression approaches such as distillation and pruning.
In this sense, our work is paving the way for interesting and novel combinations of different techniques that are potentially complementary in nature.

\section{Acknowledgements}
This work was performed under the auspices of the U.S. Department of Energy by Lawrence Livermore National Laboratory under Contract DE-AC52-07NA27344 and was supported by the LLNL-LDRD Program under Project No. 22-SI-020. LLNL-CONF-864812.






\bibliographystyle{plainnat}
\bibliography{refs}

\newpage

\newpage
\appendix

\section{Code Reproducibility} 
We plan to release our code upon acceptance of the paper.

\section{Decoupled Training with Noise Inputs} 
In this section, we explore the adaptability of our decoupled training in scenarios where the original task data is unavailable. This investigation aims to address the challenge of operating in a completely data-free environment. We employ uniformly sampled noise as input data, denoted as $X \sim U[-1,1]$. Remarkably, even in the absence of meaningful data, our proposed approach demonstrates significant performance enhancement, with improvements of approximately 2 to 3\%.

\begin{table*}[htb!]
\centering
\caption{Reconstruction performance of \textbf{Ours} ($\mathcal{L}_{KD}$ only in the second phase) with noise input data}
\label{test_accuracy_nern+decoupling_noise}
\vspace{0.2em}
\centering 
\scalebox{0.85}{
\begin{tabular}{lccc}  
\cmidrule[1.5pt](lr){1-4}
\multirow{3}{*}{\makecell{CIFAR10\vspace{5pt}\\Method (In-Filter)}} & \multirow{3}{*}{\makecell{Original\\ResNet20}} & \multicolumn{2}{c}{Hidden 140}   \\ 
\cmidrule(lr){3-4} 
 & & {Recon-only} & \textbf{Ours} \\
\midrule
Size & 1.03MB & 0.36MB & 0.36MB  \\
Acc. ($\uparrow$, \%) & 91.69\%  & 85.64\%\scriptsize{$\pm$0.39}  & \textbf{87.25\%\scriptsize{$\pm$0.02}}   \\
\cmidrule[1.5pt](lr){1-4}

\multirow{3}{*}{\makecell{STL10\vspace{5pt}\\Method (In-Filter)}} & \multirow{3}{*}{\makecell{Original\\ResNet56}} & \multicolumn{2}{c}{Hidden 320} \\ 
\cmidrule(lr){3-4} 
 & & {Recon-only} & \textbf{Ours}\\
\midrule
Size & 3.25MB & 1.48MB & 1.48MB  \\
Acc. ($\uparrow$, \%) & 76.31\%  & 72.80\%\scriptsize{$\pm$1.36} & \textbf{74.04\%\scriptsize{$\pm$0.009}}  \\
\cmidrule[1.5pt](lr){1-4}

\multirow{3}{*}{\makecell{CIFAR100\vspace{5pt}\\Method (In-Filter)}} & \multirow{3}{*}{\makecell{Original\\ResNet56}} & \multicolumn{2}{c}{Hidden 320} \\ 
\cmidrule(lr){3-4} 
 & & {Recon-only} & \textbf{Ours} \\
\midrule
Size & 3.25MB & 1.48MB & 1.48MB  \\
Acc. ($\uparrow$, \%) & 71.37\%  & 61.31\%\scriptsize{$\pm$0.45} & \textbf{64.39\%\scriptsize{$\pm$0.01}}  \\

\bottomrule
\end{tabular}}
\end{table*}

\section{Loss Landscape Analysis} 
\begin{figure}[htbp]
     \vspace{-3mm}
     \begin{subfigure}[b]{0.32\textwidth}
         \includegraphics[width=\textwidth]{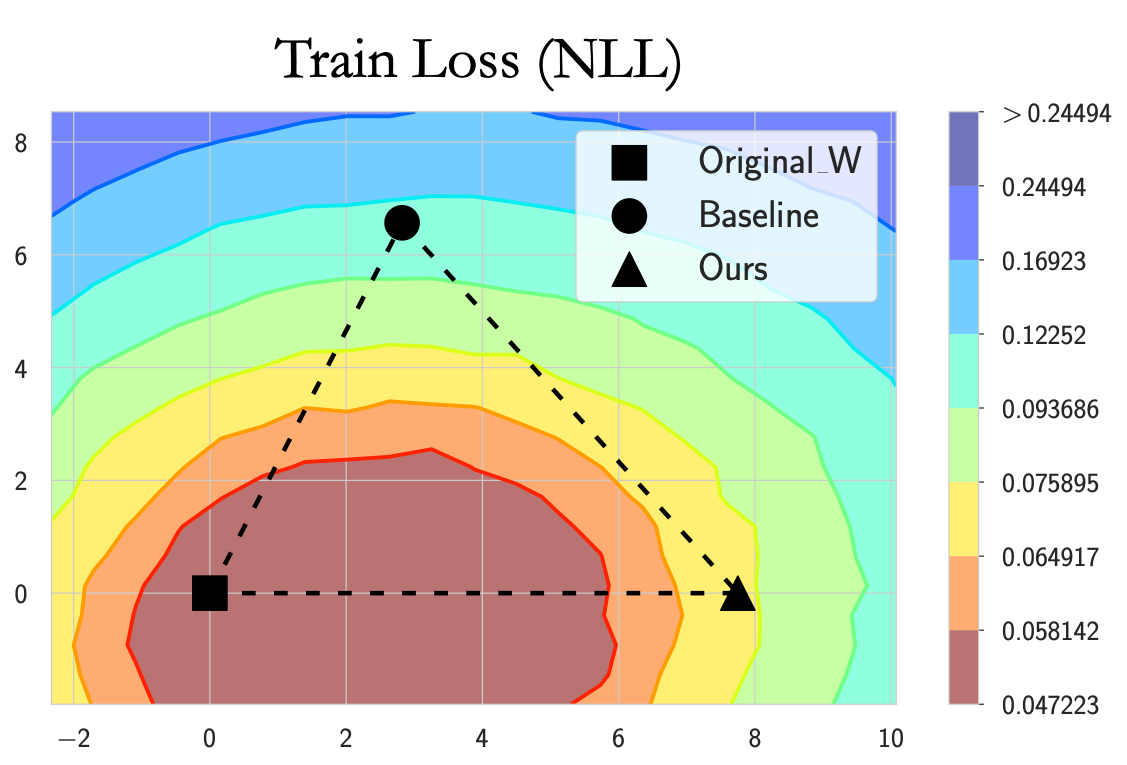}
     \end{subfigure}
     \begin{subfigure}[b]{0.32\textwidth}
         \includegraphics[width=\textwidth]{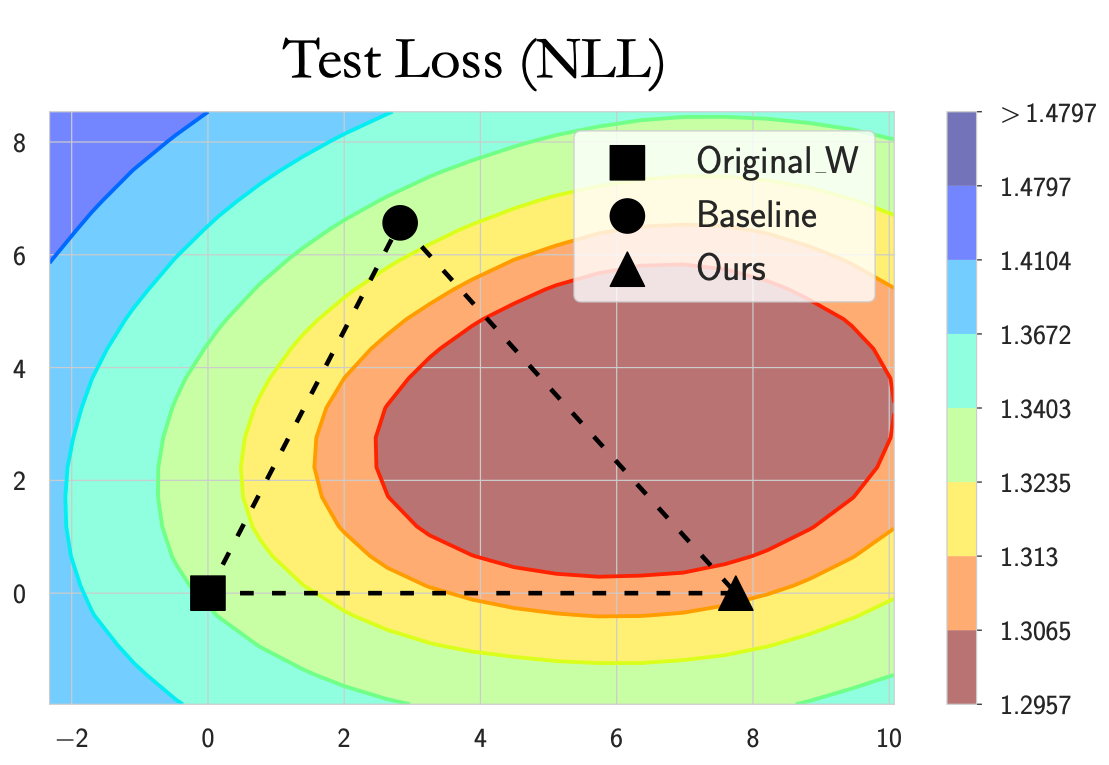}
     \end{subfigure}
     \begin{subfigure}[b]{0.32\textwidth}
         \includegraphics[width=\textwidth]{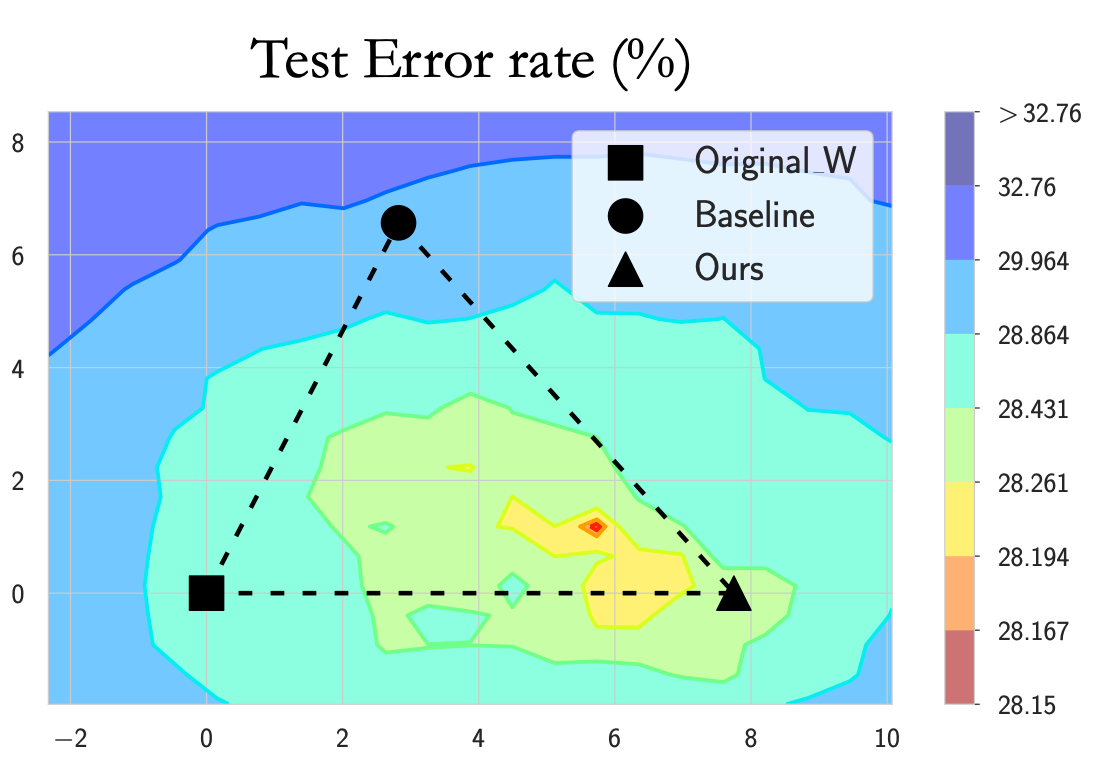}
     \end{subfigure}
     \vspace{0mm}
     \caption{Loss landscape analysis with comparisons among weights from the original networks, Baseline ($\mathcal{L}_{recon}+\mathcal{L}_{KD}+\mathcal{L}_{FMD}$), and \textbf{Ours} ($\mathcal{L}_{KD}$ only in the second phase)}
     \label{fig:loss_landscape_decoupled}
     \vspace{-3mm}
 \end{figure}

Understanding the landscape of the loss function in the weight space provides valuable insights into the optimization process and the behavior of neural networks. Here, we conduct a comprehensive loss landscape analysis to compare the learned weights from each method. As shown in Figure \ref{fig:loss_landscape_decoupled}, the weights found by our decoupled training lie on the periphery of the most desirable solutions. This suggests that decoupled training enables the predictor to explore better optima by exclusively learning from the predictive knowledge of the original network. Additionally, when examining the gap between train loss and test loss, our method exhibits a smaller gap, aligning with the findings in \citep{keskar2016large}. This indicates that our approach not only finds better optima but also generalizes well to unseen data.

\begin{figure}[htbp]
     \vspace{-3mm}
     \begin{subfigure}[b]{0.32\textwidth}
         \includegraphics[width=\textwidth]{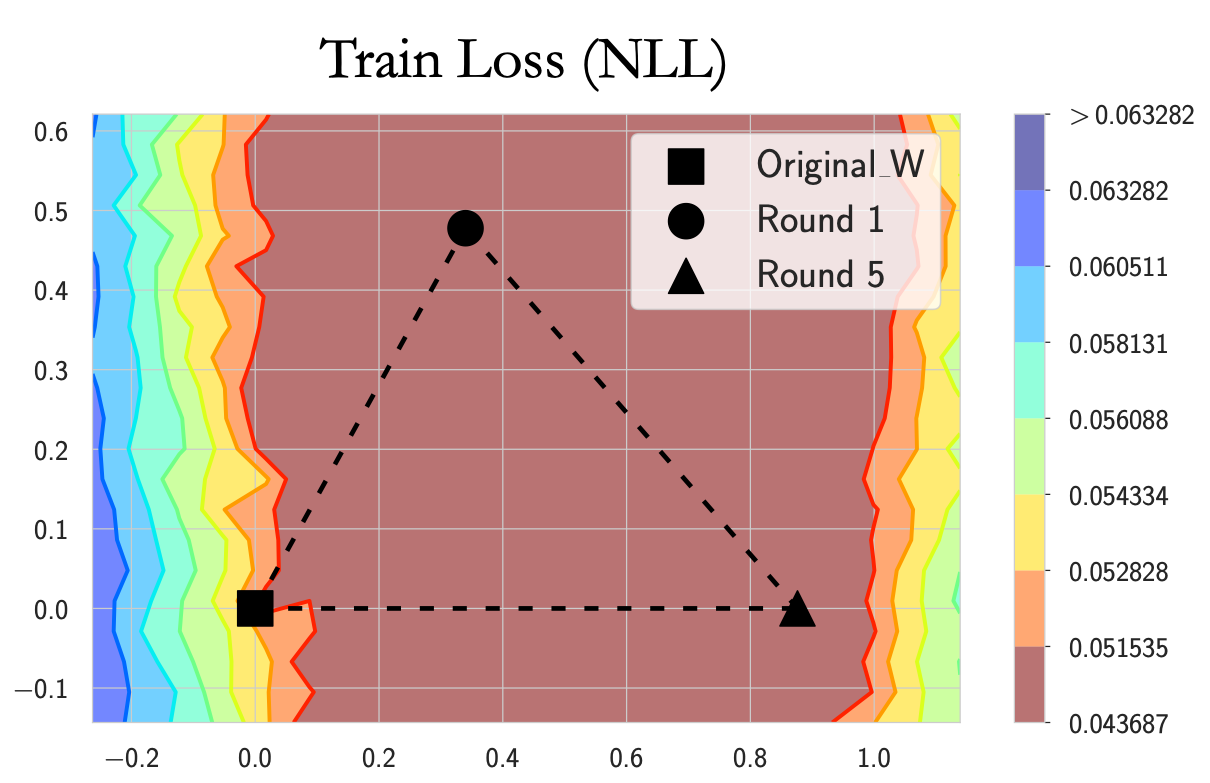}
     \end{subfigure}
     \begin{subfigure}[b]{0.32\textwidth}
         \includegraphics[width=\textwidth]{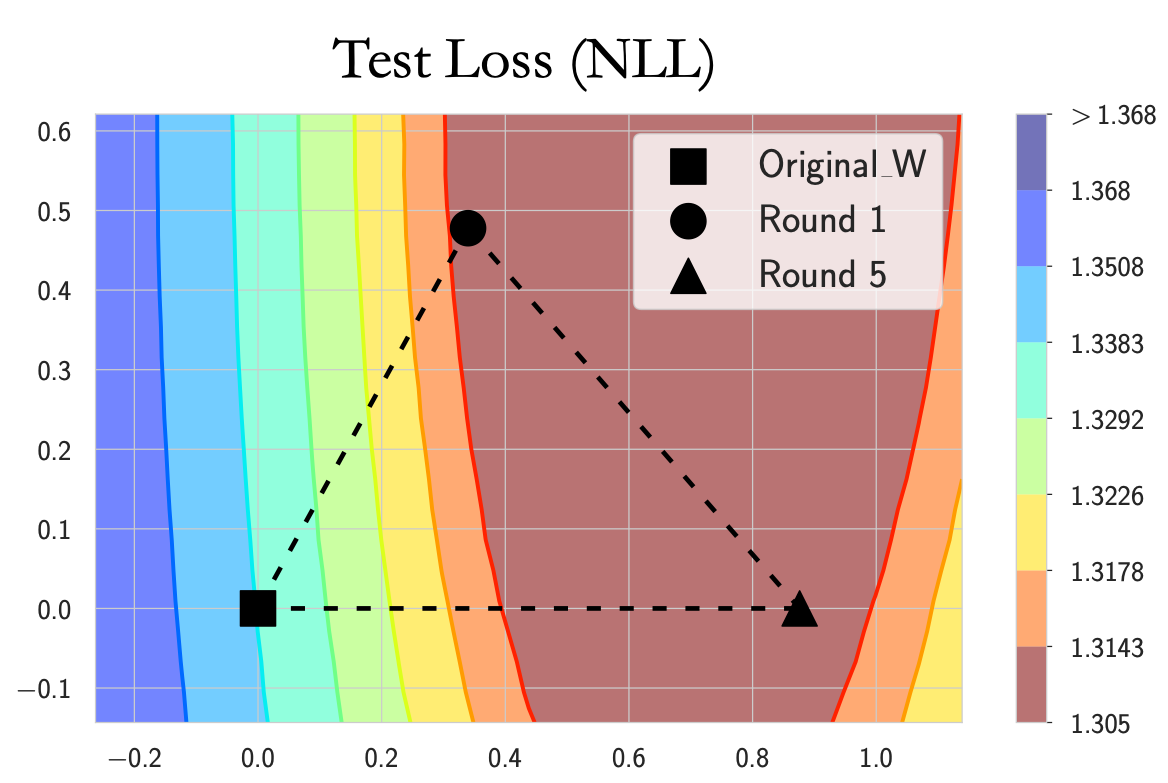}
     \end{subfigure}
     \begin{subfigure}[b]{0.32\textwidth}
         \includegraphics[width=\textwidth]{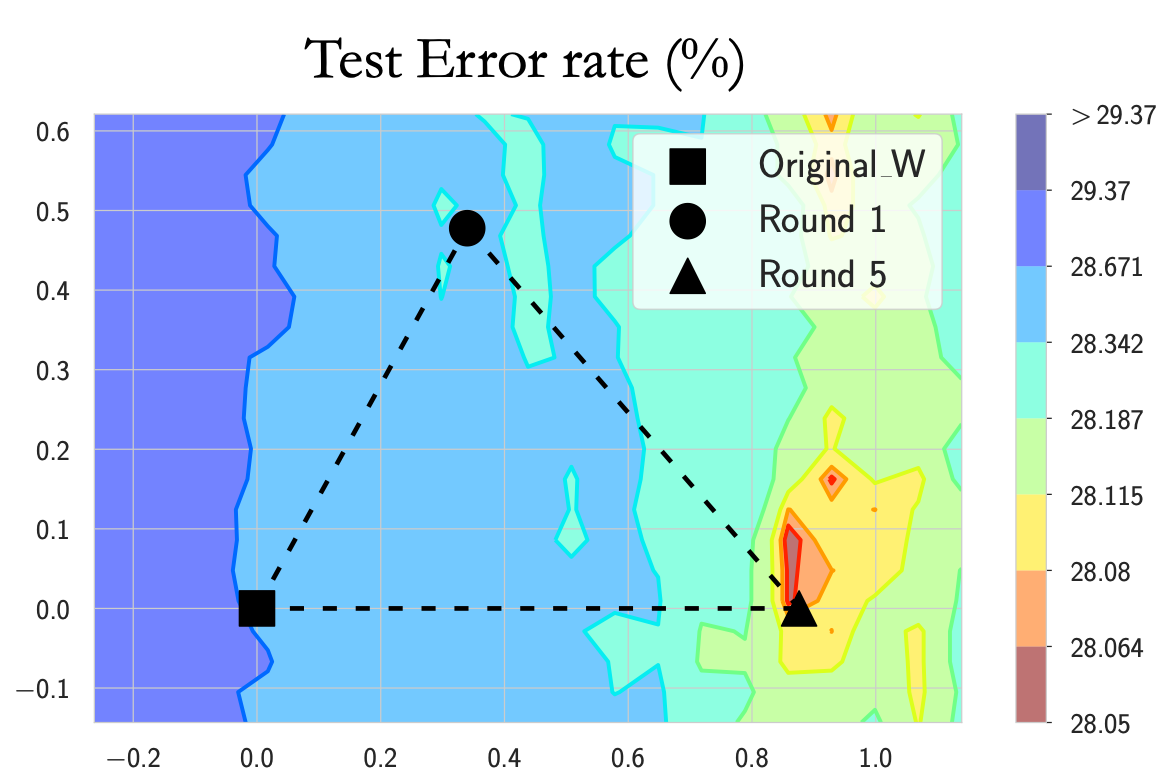}
     \end{subfigure}
     \vspace{0mm}
     \caption{Loss landscape analysis comparing weights from the original network with those from the reconstructed models in the first and last rounds.}
     \label{fig:loss_landscape_inception}
     \vspace{-3mm}
 \end{figure}

We also conduct the same analysis for Section \ref{sec:NeRN_Inception_Training}. Here, we visualize the loss landscape by interpolating the weights among the original network, and the first and last rounds of the reconstructed model. As shown in Figure \ref{fig:loss_landscape_inception}, all three weights belong to the same local extrema in the loss landscape. This observation is expected since the reconstruction loss constrains all weights to be close to the original model. For the testing loss/error, the reconstruction process appears to enhance the generalization performance.

\section{Datasets and Training Details}\label{detailed_info_training} 
\subsection{Dataset Description}
\begin{itemize}
    \item \textbf{CIFAR-10 \citep{krizhevsky2009learning}:} The CIFAR-10 dataset consists of 60,000 32x32 color images in 10 classes, with 6,000 images per class. The dataset is divided into 50,000 training images and 10,000 testing images. Each image is labeled with one of the following classes: airplane, automobile, bird, cat, deer, dog, frog, horse, ship, or truck.
    
    \item \textbf{CIFAR-100 \citep{krizhevsky2009learning}:} Similar to CIFAR-10, the CIFAR-100 dataset contains 60,000 32x32 color images, but organized into 100 classes, with 600 images per class. The dataset is divided into 50,000 training images and 10,000 testing images. Each image is labeled with one of the 100 fine-grained classes, which are grouped into 20 coarse-grained superclasses.
    
    \item \textbf{STL-10 \citep{coates2011analysis}:} The STL-10 dataset comprises 10,000 labeled 96x96 color images, with 5,000 images for training and 5,000 for testing. The dataset contains images from 10 different classes: airplane, bird, car, cat, deer, dog, horse, monkey, ship, and truck. Unlike CIFAR, STL-10 also includes a pre-defined unlabeled dataset for unsupervised learning tasks.
    
    \item \textbf{ImageNet-1K \citep{deng2009imagenet}:} ImageNet-1K is a large-scale dataset consisting of over 1.2 million high-resolution images across 1,000 different classes. It is widely used for image classification, object detection, and other computer vision tasks. The dataset is divided into training (1.28 million images), validation (50,000 images), and test sets (100,000 images). Each image is labeled with one of the 1,000 object categories.
\end{itemize}

\subsection{Training Details}
\noindent\textbf{Training of Baseline: } In baseline training, we follow the same settings outlined in \citep{ashkenazi2022nern}. The Baseline method employs a Multi-layer Perceptron (MLP) with $5$ layers as a predictor, with varying hidden sizes. Training is conducted using the ranger optimizer \citep{wright2019ranger} with a learning rate of $5e-3$. The number of epochs for training is $350$ for CIFAR-10 and STL-10, $450$ for CIFAR-100, and $16\times 10^{4}$ iterations for ImageNet experiments. Similar to minibatch sampling in standard stochastic optimization, during each training step of Baseline, it predicts all reconstructed weights but optimizes only on a mini-batch of them. The weights batch method employed is a random weighted batch, using weighted sampling with a probability of $1-p_{uni}$, where $p_{uni}=0.8$, and a batch size of $4096$ for CIFAR-10, CIFAR-100, and STL-10 datasets. For ImageNet, the experiment was conducted with a minibatch size of $2^{16}$. Hyperparameters $\alpha$ and $\beta$  in learning objectives are set to $1e-5$ for CIFAR-100 and STL-10 datasets, and to $1e-4$ to CIFAR-10, and $1e-6$ for ImageNet. Based on empirical observations, increasing the hyperparameter values during training to emphasize the distillation process causes the Baseline method to experience highly unstable training, often resulting in convergence failure. Therefore, we opted to use the same values as suggested by the authors. 

When training predictors, there are two types of permutation-based smoothness: In-Filter and Cross-Filter. Both approaches do not show significant difference in terms of accuracy. The order of weights in the original network remains unchanged; this smoothness only affects the order in which the predictor processes the kernels. In all experiments, In-Filter smoothness was used for CIFAR-10, CIFAR-100, and STL-10 datasets, while Cross-Filter smoothness was employed for the ImageNet dataset.

\noindent\textbf{Training of Progressive-Reconstruction Training: }\label{sec:appendix_progressive} We adhere to the same settings as full training in the Baseline method, including the number of epochs, batch size, learning rate, and other parameters. To isolate and illustrate the reconstruction's pure effect, the predictor is trained only with the reconstruction loss, $\mathcal{L}_{recon}$. For the next round of progressive-reconstruction training, we select the best-performing models from the previously reconstructed network, determined across three trials with different random seeds, as the target network. If the performance does not surpass that of the target network, we conclude the round. To elucidate the training protocols, we present the results of all three trials conducted on the CIFAR-100 dataset. As observed in the trend of improvement, the gap in enhancement diminishes as the rounds progress.

\begin{table*}[htb!]
\centering
\caption{Evaluation performance of large predictor networks via progressive reconstruction. We report all results in three trials. The colored box represents the target performance for the next round. }
\label{result_progressive_all_trials}
\vspace{0.2em}
\centering 
\scalebox{0.80}{
\begin{tabular}{lccccccc}  
\cmidrule[1.5pt](lr){1-8}
\multirow{2}{*}{\makecell{CIFAR100\vspace{5pt}\\Method}} & \# Trial & \multirow{2}{*}{\makecell{Original\\ResNet56}} & \multicolumn{5}{c}{Hidden 680 (\textsl{CR}$>$1)} \\ 
\cmidrule(lr){4-8} & & & \textit{Round 1} & \textit{Round 2} & \textit{Round 3} & \textit{Round 4} & \textit{Round 5}\\
\midrule
Accuracy ($\uparrow$, \%) & 1 & \colorbox{blue!20}{71.37}  & 71.62 & 71.73 & 71.80 & 71.91 & 71.92 \\
Accuracy ($\uparrow$, \%) & 2 & -  & 71.59 & 71.69  & \colorbox{blue!20}{71.89} & \colorbox{blue!20}{71.96} & 72.05 \\
Accuracy ($\uparrow$, \%) & 3 & -  & \colorbox{blue!20}{71.63} & \colorbox{blue!20}{71.79} & 71.86 & 71.91 & 71.99 \\
\midrule
mean$\pm$std & & & \textbf{71.61\%\scriptsize{$\pm$0.01}} & \textbf{71.73\%\scriptsize{$\pm$0.04}} & \textbf{71.85\%\scriptsize{$\pm$0.03}} & \textbf{71.92\%\scriptsize{$\pm$0.02}} & \textbf{71.98\%\scriptsize{$\pm$0.05}}\\
\bottomrule
\end{tabular}}
\end{table*}

\noindent\textbf{Training of Decoupled Training: } For our proposed decoupled training, we initiate training from the solution obtained by the best-performing model in Recon-only models, as described in Section \ref{sec:appendix_progressive}. As discussed in Section \ref{sec:NeRN_Decoupled_Training}, fully separating the two phases provides the advantage of enabling more flexible learning during the fine-tuning stage of the second phase. This flexibility allows for the exploration of different optimizers compared to those used in the first phase. Moreover, to maximize the performance of the predictor network, it enables the application of various advanced distillation methods that have been previously explored.

In this experiment, we fine-tune predictors in the second phase for 100 epochs for CIFAR-10, CIFAR-100, and STL-10, and $10^5$ iterations for ImageNet, but even with a much smaller number of epochs/iterations, we observe comparable performance. We employ either Adam \citep{kingma2014adam} or Ranger \citep{wright2019ranger} optimizers, and in most cases, both yield similar performance. Additionally, in the second phase with $\mathcal{L}_{KD}$, we empirically observed that adding weights to the $\mathcal{L}_{KD}$ with $\alpha<1$ sometimes helps improve convergence, resulting in better performance. Therefore, in most experiments, we set $\alpha$ to $0.01$.

\section{Weight Frequency Component Comparison}
\begin{figure}[htbp]
    \vspace{-4mm}
    \begin{subfigure}[b]{0.50\textwidth}
        \includegraphics[width=\textwidth]{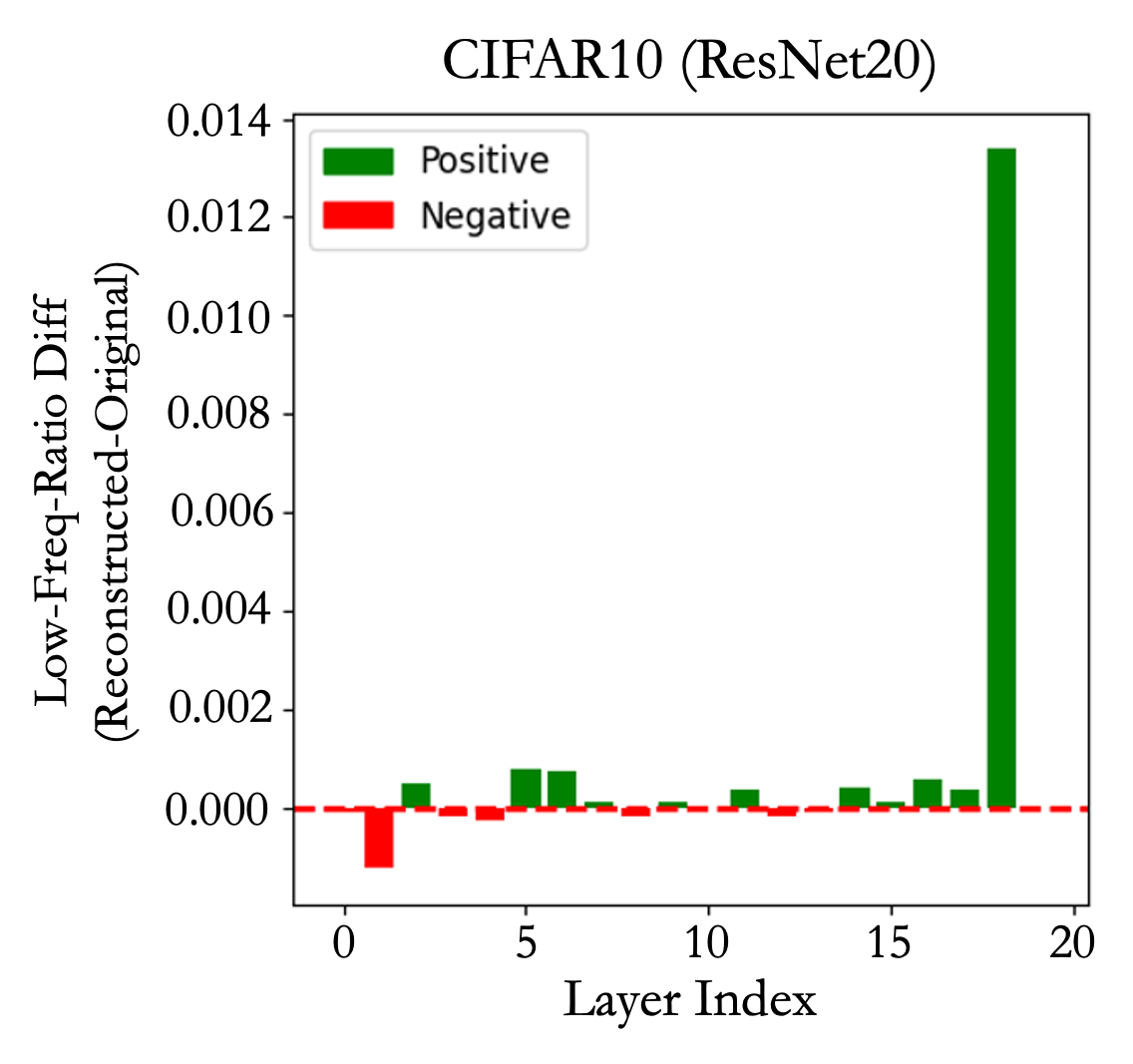}
    \end{subfigure}
    \begin{subfigure}[b]{0.50\textwidth}
        \includegraphics[width=\textwidth]{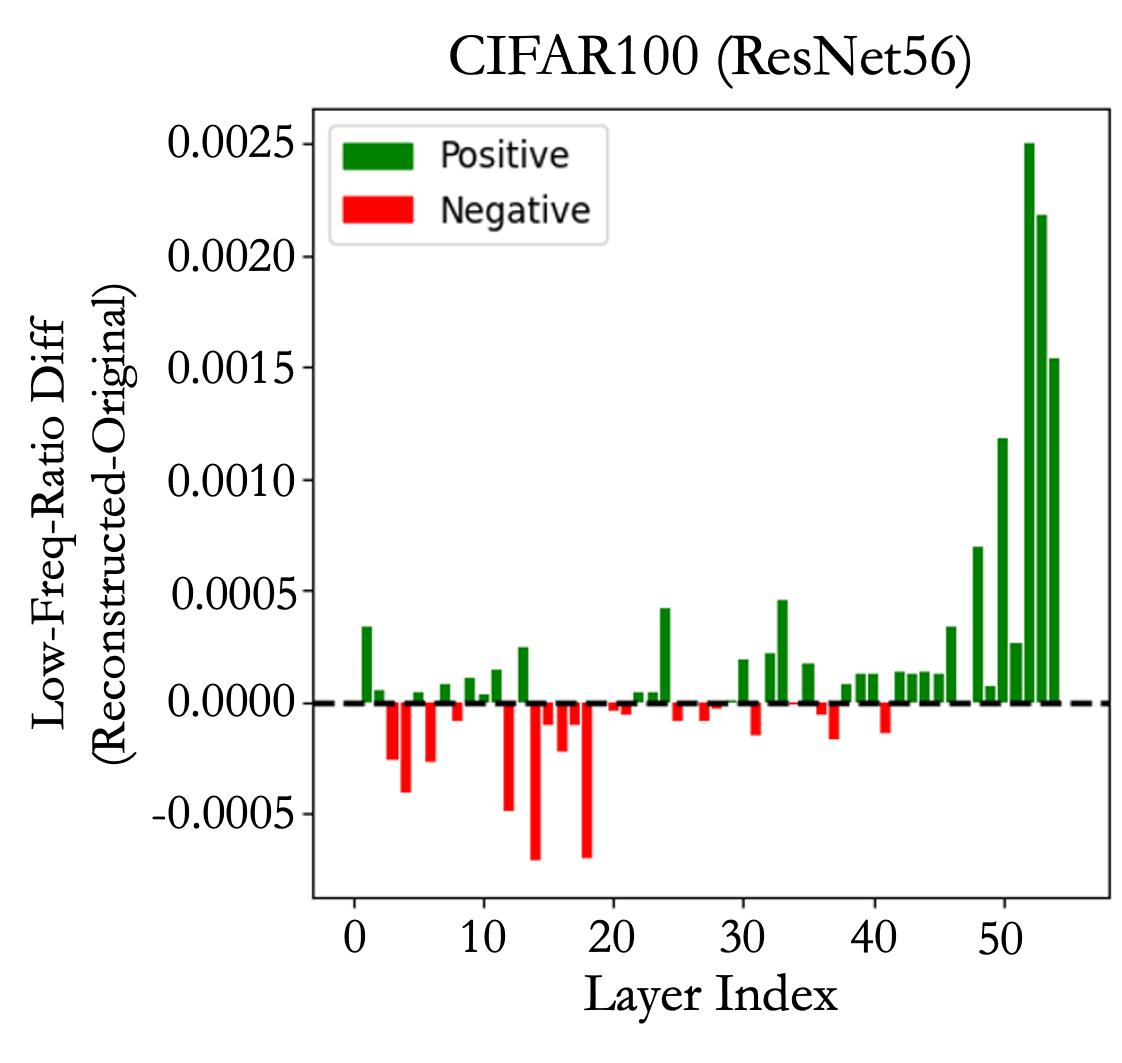}
    \end{subfigure}
    \begin{subfigure}[b]{0.50\textwidth}
        \includegraphics[width=\textwidth]{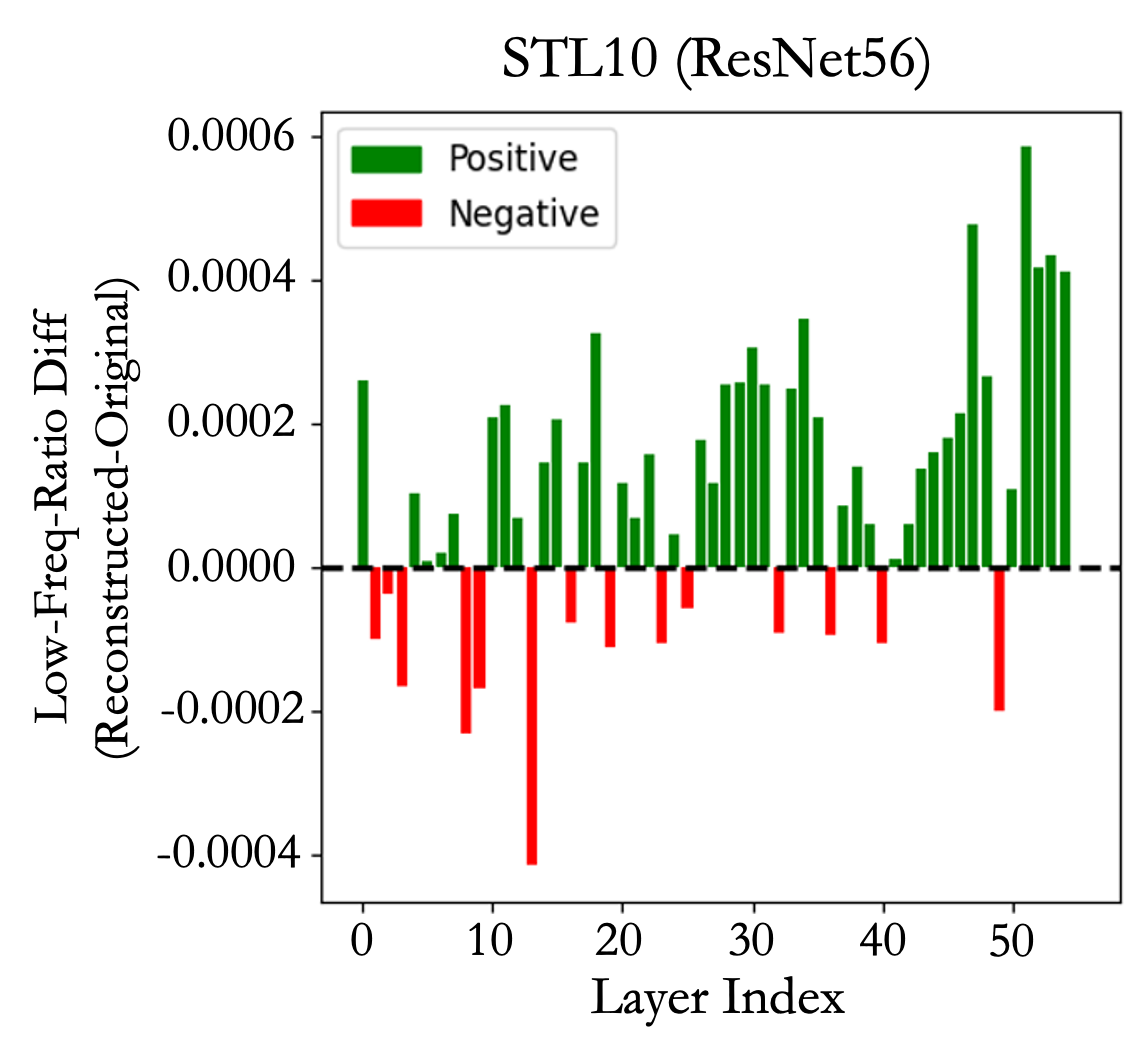}
    \end{subfigure}
    \caption{The per-layer of lower frequency component comparison between the original network weight and the reconstructed weight found in the last round of progressive-reconstruction.}
    \label{fig:freq_ratio_appendix}
\end{figure}

As observed in the low-frequency analysis, we extend our investigation to the solution obtained in the final round of progressive-reconstruction. The ratio (on the $y$-axis) in this analysis measures the proportion of the total variance (energy) of the matrix \(\mathbf{M}\) captured by the first half of the singular values. A higher ratio signifies that the lower frequency components are more dominant. In Figure \ref{fig:freq_ratio_appendix}, we present a per-layer comparison of this ratio between the reconstructed and original weights. Despite the presence of noise in the pattern, a clear trend emerges: the reconstructed weights generally exhibit more dominant lower frequency components compared to the original weights, with more pronounced changes occurring in the later layers. Intriguingly, this unexpected finding aligns with observations on rank reduction to enhance the reasoning capabilities of large language models \citep{sharma2023truth}, suggesting a broader implication regarding the relationship between the frequency components of weights and model generalizability.

\section{Interpolation Between Two Extremes}
In addition to investigating frequency analysis, we further examine individual pairs of weights by directly interpolating between the original weight $w_{o}$ and each round's solution from the progressive-reconstruction $\bar{w}$ to gain insight into the relationship between reconstruction error and model accuracy.

\begin{figure}[htbp]
    \begin{subfigure}[b]{0.50\textwidth}
        \includegraphics[width=\textwidth]{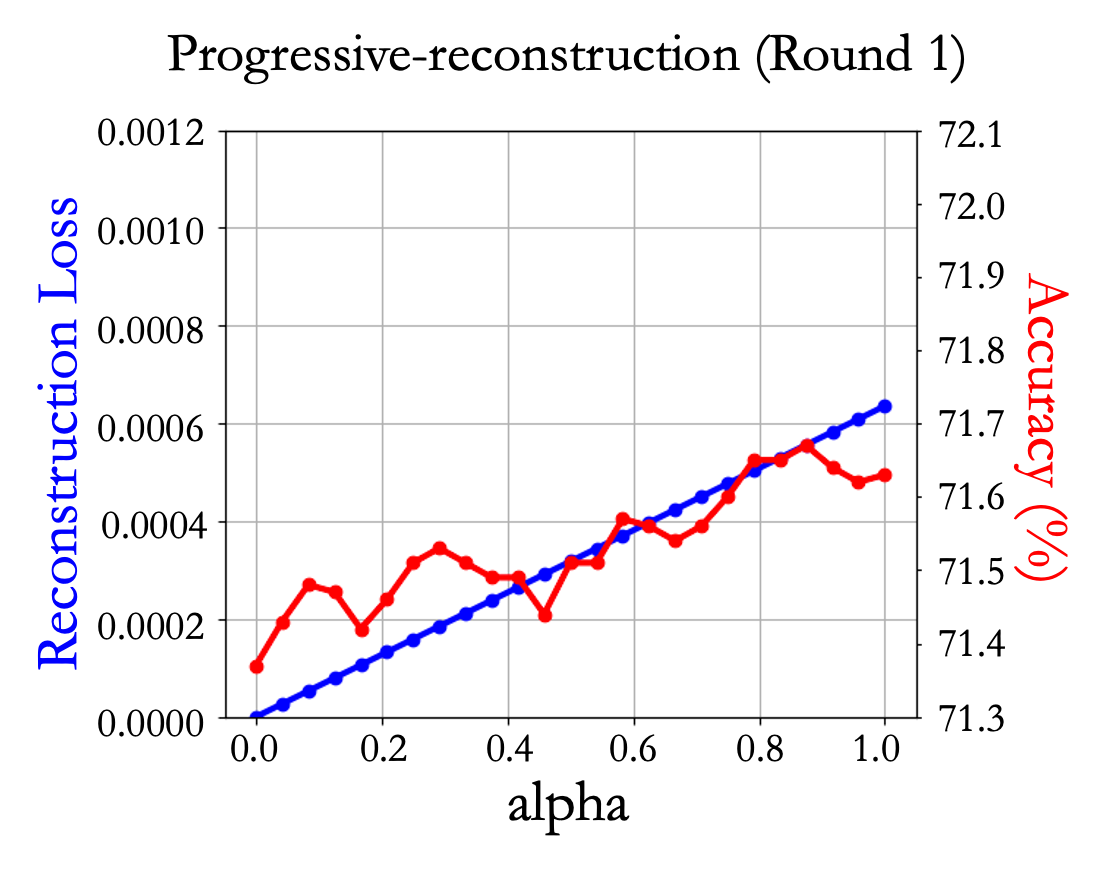}
    \end{subfigure}
    \begin{subfigure}[b]{0.50\textwidth}
       \includegraphics[width=\textwidth]{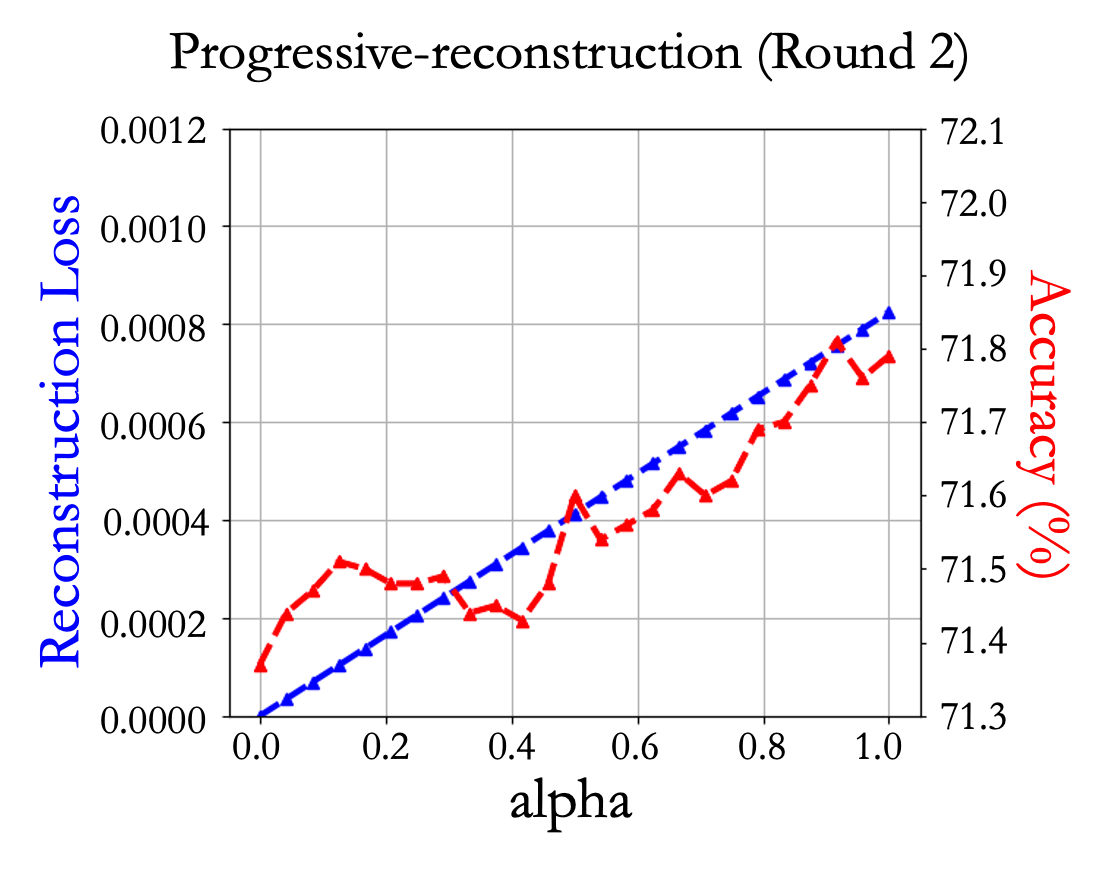}
    \end{subfigure}
    \begin{subfigure}[b]{0.50\textwidth}
       \includegraphics[width=\textwidth]{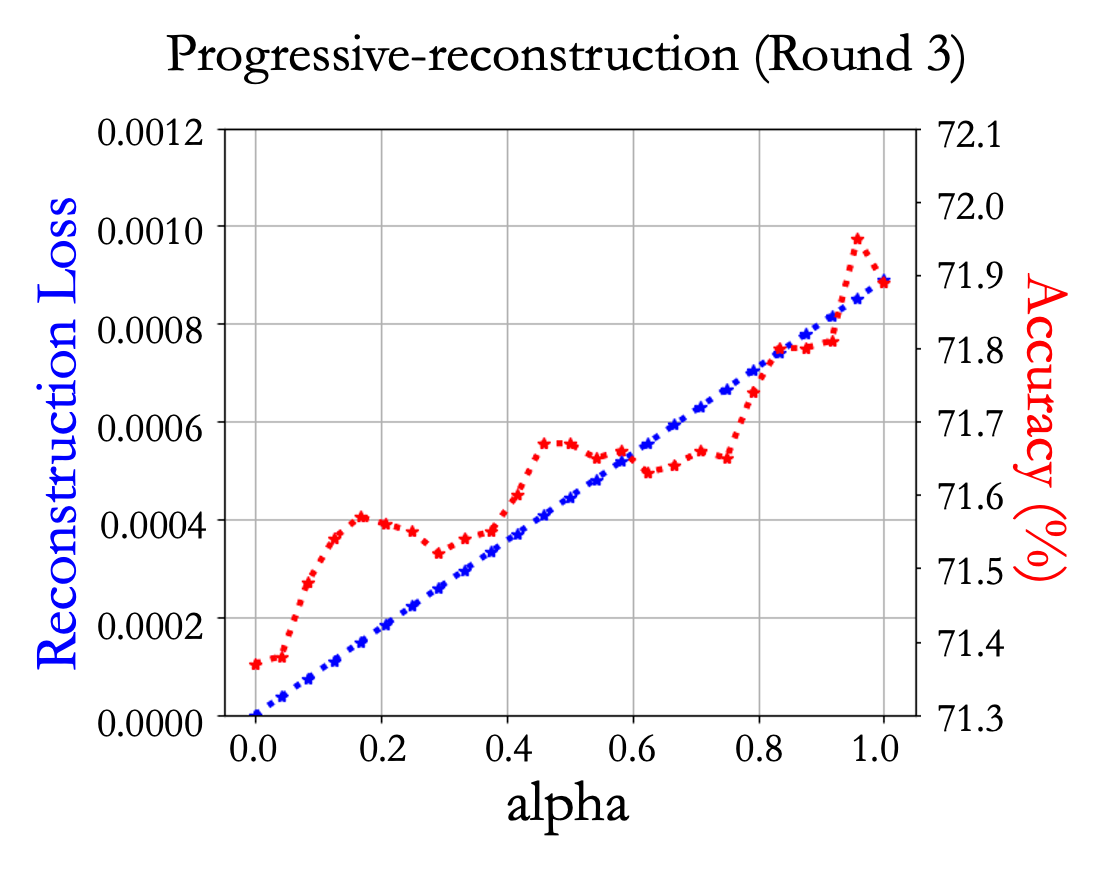}
    \end{subfigure}    
    \begin{subfigure}[b]{0.50\textwidth}
       \includegraphics[width=\textwidth]{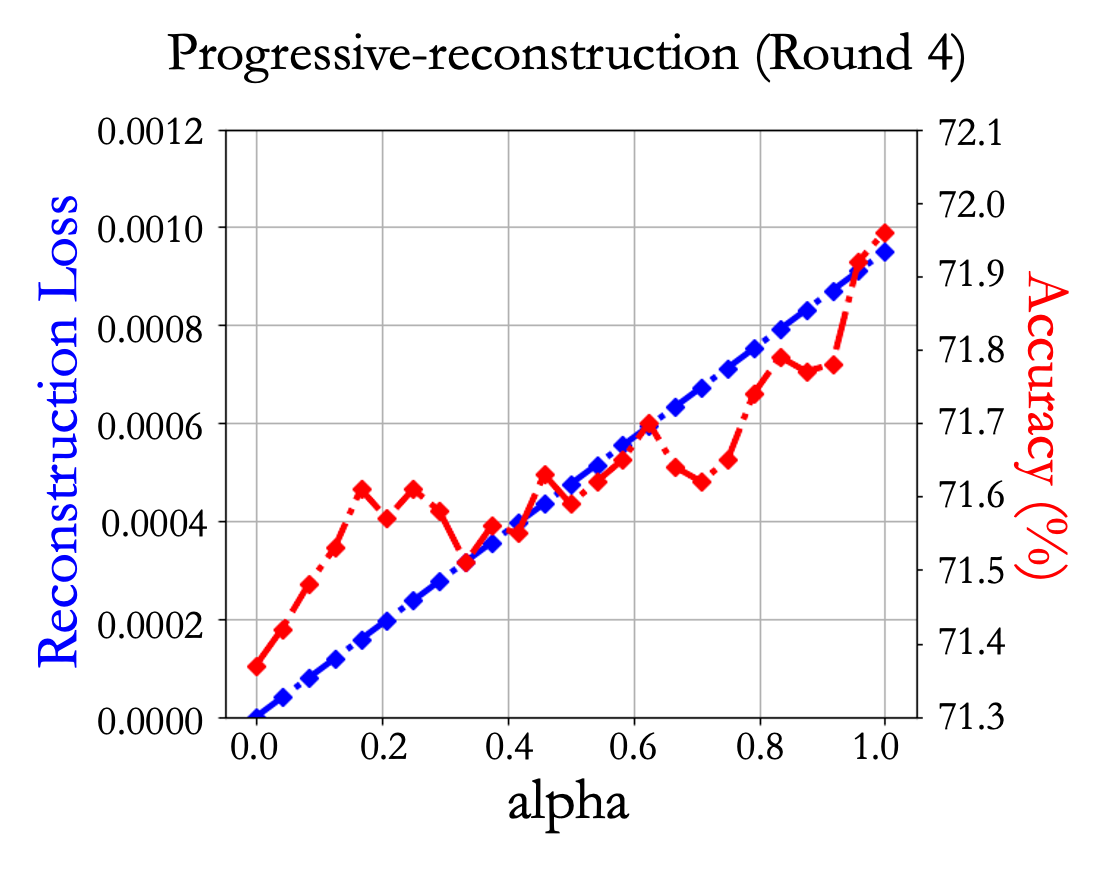}
    \end{subfigure}    
    \begin{subfigure}[b]{0.50\textwidth}
        \includegraphics[width=\textwidth]{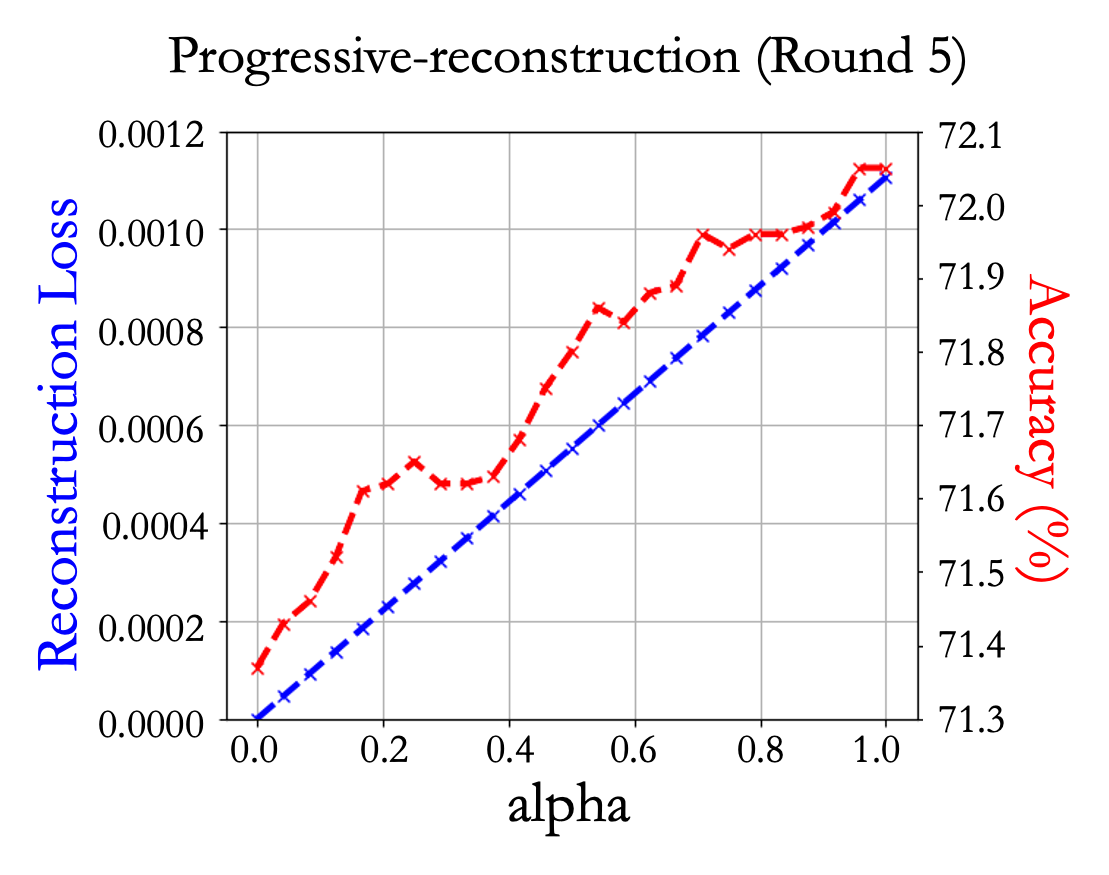}
    \end{subfigure}    

    \caption{Analysis of interpolation in a 1-D parameter space between the original weight and the solution in each round of progressive-reconstruction.}\label{fig:interpolation_1d_space_appendix}
\end{figure}

Let $w_{o}$ and $\bar{w}_{i}$ represent the original weights and the learned weights at the $i$-th round, respectively. For each value of $\alpha$ in the range $[0, 1.0]$, we generate plots showing the test accuracy (on the right $y$-axis) and the corresponding reconstruction error (on the left $y$-axis) for the interpolated weights $f((1-\alpha)w_{o}+\alpha \bar{w}_{i})$. These plots illustrate the performance at intermediate points across different values of $\alpha$, as depicted in Figure \ref{fig:interpolation_1d_space_appendix}. The plots reveal a gradual increase in both distance and accuracy as the rounds progress.

\end{document}